\documentclass[10pt,twocolumn,letterpaper]{article}

\usepackage{times}
\usepackage{epsfig}
\usepackage{graphicx}
\usepackage{amsmath}
\usepackage{amssymb}
\usepackage{multirow}
\usepackage[vlined,ruled]{algorithm2e}
\usepackage{xcolor}
\usepackage{tabularx}
\usepackage{makecell}
\usepackage{soul}

\usepackage{tikz}

\newcommand{\Best}[1]{\textbf{\textcolor{black}{#1}}}
\newcommand{\B}[1]{\Best{#1}}
\newcommand{\SB}[1]{\textbf{\color{brown}{#1}}}

\makeatletter
\DeclareRobustCommand\onedot{\futurelet\@let@token\@onedot}
\def\@onedot{\ifx\@let@token.\else.\null\fi\xspace}

\def\etal{\emph{et al}\onedot}
\makeatother

\usepackage[pagebackref=true,breaklinks=true,letterpaper=true,colorlinks,bookmarks=false]{hyperref}

\date{}

\begin{document}

\title{Non-Local Video Denoising by CNN}
\author{
\begin{tabularx}{\linewidth}{X}
\hfill \makecell{Axel Davy} \hfill \makecell{Thibaud Ehret} \hfill \makecell{Jean-Michel Morel}  \hfill\null\\
\hfill \makecell{Pablo Arias} \hfill \makecell{Gabriele Facciolo} \hfill\null\\
\end{tabularx} \\
CMLA, ENS Cachan, CNRS\\
Universit\'e Paris-Saclay, 94235 Cachan, France\\
{\tt\small axel.davy@ens-cachan.fr}\thanks{Work supported by IDEX  Paris-Saclay IDI 2016, ANR-11-IDEX-0003-02, ONR  grant N00014-17-1-2552,  CNES MISS project, DGA Astrid ANR-17-ASTR-0013-01, DGA ANR-16-DEFA-0004-01. The TITAN V used for this research was donated by the NVIDIA Corporation.}
}

\maketitle

\begin{abstract}
Non-local patch based methods were until recently state-of-the-art for image denoising but are now outperformed by CNNs. Yet they are still the state-of-the-art for video denoising, as video redundancy is a key factor to attain high denoising performance. The problem is that CNN architectures are hardly compatible with the search for self-similarities.
In this work we propose a new and efficient way to feed video self-similarities to a CNN. The non-locality is incorporated into the network via a first non-trainable layer which finds for each patch in the input image its most similar patches in a search region. The central values of these patches are then gathered in a feature vector which is assigned to each image pixel. This information is presented to a CNN which is trained to predict the clean image. We apply the proposed architecture to image and video denoising. For the latter patches are searched for in a 3D spatio-temporal volume. The proposed architecture achieves state-of-the-art results. To the best of our knowledge, this is the first successful application of a CNN to video denoising.
\end{abstract}

\section{Introduction}

Advances in image sensor hardware have steadily improved the acquisition quality of image and video cameras. However, a low signal-to-noise ratio is unavoidable in low lighting conditions if the exposure time is limited (for example to avoid motion blur). This results in high levels of noise, which negatively affects the visual quality of the video and hinders its use for many applications.
As a consequence, denoising is a crucial component of any camera pipeline.
Furthermore, by interpreting denoising algorithms as proximal operators, several inverse problems in image processing can be solved by iteratively applying a denoising algorithm \cite{17-romano-red}. Hence  the need for video denoising algorithms with a low running time.

\paragraph{Literature review on image denoising.}

Image denoising has a vast literature where a variety of methods have been applied: PDEs and variational methods (including MRF models) \cite{Rudin1992-ROF,Caselles2010-intro-tv,Roth2005-foe}, transform domain methods \cite{Donoho1994}, non-local (or patch-based) methods \cite{Buades2005,dabov2007image}, multiscale approaches \cite{Facciolo2017-multiscaler}, etc. See \cite{Lebrun2012} for a review. In the last two or three years, CNNs have taken over the state-of-the-art. In addition to attaining better results, CNNs are amenable to efficient parallelization on GPUs potentially enabling real-time performance.
We can distinguish two types of CNN approaches: \emph{trainable inference networks} and \emph{black box} networks.

In the first type, the architecture mimics the operations performed by a few iterations of optimization algorithms used for MAP inference with MRFs prior models. Some approaches are based on the Field-of-Experts model \cite{05-roth-foe}, such as \cite{09-barbu-arf,14-schmidt-csf,Chen2017TrainableRestoration}. The architecture of \cite{16-vemulapalli-deep-gcrf} is based on EPLL \cite{11-zoran-epll}, which models the \textit{a priori} distribution of image patches as a Gaussian mixture model.
Trainable inference networks reflect the operations of an optimization algorithm, which leads in some cases to unusual architectures, and to some restrictions in the network design. For example, in the \emph{trainable reaction diffusion network} (TRDN) of \cite{Chen2017TrainableRestoration} even layers must be an image (i.e. have only one feature). As pointed out in \cite{17-kobler-variational-nets} these architectures have strong similarities with the residual networks of \cite{He2016}.

The black-box approaches treat denoising as a standard regression problem. They do not use much of the domain knowledge acquired during decades of research in denoising. In spite of this, these techniques are currently topping the list of state-of-the-art algorithms.
The first denoising approaches using neural networks were proposed in the mid and late 2000s. Jain and Seung \cite{09-jain-denoising-cnn} proposed a five layer CNN with $5\times5$ filters, with 24 features in the hidden layers and sigmoid activation functions. Burger et al. \cite{12-burger-mlp} reported the first state-of-the-art results with a multilayer perceptron trained to denoise $17\times 17$ patches, but  with a heavy architecture. More recently, DnCNN~\cite{Zhang2017BeyondDenoising} obtained impressive results with a far lighter 17 layer deep CNN with $3\times 3$ convolutions, ReLU activations and batch normalization \cite{Ioffe2015BatchShift}.  This work also proposes a blind denoising network that can denoise an image with an unknown noise level $\sigma\in[0,55]$, and a multi-noise network trained to denoise blindly three types of noise. A faster version of DnCNN, named FFDNet, was proposed in \cite{17-zhang-ffdnet}, which also allows handling noise with spatially variant variance $\sigma(x)$ by adding the noise variance map as an additional input. The architectures of DnCNN and FFDnet keep the same image size throughout the network. Other architectures~\cite{16-mao-red,16-santhanam-rbdn,18-chen-see-in-the-dark} use pooling or strided convolutions to downscale the image, and then up-convolutional layers to upscale it back. Skip connections connect the layers before the pooling with the output of the up-convolution to avoid loss of spatial resolution. Skip connections are used extensively in~\cite{17-tai-memnet}.

Although these architectures produce very good results, for textures formed by repetitive patterns, non-local patch-based methods still perform better \cite{Zhang2017BeyondDenoising,12-burger-mlp}. Some works have therefore attempted to incorporate the non-local patch similarity into a CNN framework. Qiao \etal \cite{17-qiao-tnlrd} proposed inference networks derived from the non-local FoE MRF model \cite{11-sun-nlfoe}. This can be seen as a non-local version of the TRDN network of \cite{Chen2017TrainableRestoration}. A different non-local TRDN was introduced by \cite{17-lefkimmiatis-nlcnn}. BM3D-net \cite{18-yang-bm3dnet} pre-computes for each pixel a stack of similar patches which are fed into a CNN, which reproduces the operations done by (the first step of) the BM3D algorithm: a linear transformation of the group of patches, a non-linear shrinkage function and a second linear transform (the inverse of the first). The authors train the linear transformations and the shrinkage function. In~\cite{18-cruz-nn3} the authors propose an iterative approach that can be used to reinforce non-locality to any denoiser. Each iteration consists of the application of the denoiser followed by a non-local filtering step using a fixed image (denoised with BM3D) for computing the non-local correspondences. This approach obtains good results and can be applied to any denoising network. An inconvenience is that the resulting algorithm requires to iterate the denoising network.
Trainable non-local modules have been proposed recently by using differentiable relaxations of the 1 nearest neighbors \cite{Liu2018-nl-rnns} and $k$ nearest neighbors \cite{Plotz2018-NNN} selection rules.

\paragraph{Literature review on video denoising.}

CNNs have been successfully applied to several video processing tasks such as deblurring~\cite{su2017deep}, video frame synthesis~\cite{17-liu-frame-synth} or super-resolution~\cite{15-huang-multi-frame-superres,18-sajjadi-frame-recurrent-video-superres}, but their application to video denoising has been limited so far. In~\cite{chen2016deep} a recurrent architecture is proposed, but the results are below the state-of-the-art. Some works have tackled the related problem of burst denoising. Recently \cite{godard2017deep,18-mildenhall-kpn} focused on the related problem of image burst denoising reporting very good results.

In terms of output quality the state-of-the-art is achieved by patch-based methods~\cite{Dabov2007v,Maggioni2012,Arias2018,ehret2017global,buades2016patch,wen2017joint}. They exploit drastically the self-similarity of natural images and videos, namely the  fact that most patches have several similar patches around them (spatially and temporally). Each patch is denoised using these similar patches, which are searched for in a region around it. The search region generally is a space-time cube, but more sophisticated search strategies involving optical flow have also been used. Because of the use of such broad search neighborhoods these methods are called \emph{non-local}. 
While these video denoising algorithms perform very well, they often are computationally costly. Because of their complexity they are usually unfit for high resolution video processing. 

Patch-based methods usually follow three steps that can be iterated: (1) search for similar patches, (2) denoise the group of similar patches, (3) aggregate the denoised patches to form the denoised frame. %
VBM3D~\cite{Dabov2007v} improves the image denoising algorithm BM3D~\cite{dabov2007image} by searching for similar patches in neighboring frames using a ``predictive search" strategy which speeds up the search and gives some temporal consistency. %
VBM4D \cite{Maggioni2012} generalizes this idea to 3D patches.
In VNLB~\cite{Arias2015} spatio-temporal patches that were not motion compensated are used to improve the temporal consistency. In~\cite{ehret2017global} a generic search method extends every patch-based denoising algorithm into a global video denoising algorithm by extending the patch search to the entire video. SPTWO~\cite{buades2016patch} and DDVD~\cite{buades2017ddvd} use optical flow to warp the neighboring frames to each target frame. %
Each patch of the target frame is then denoised using the similar patches in this volume with a Bayesian strategy similar to~\cite{lebrun2013nonlocal}. Recently,~\cite{wen2017joint} proposed to learn an adaptive optimal transform using batches of frames.

Patch-based approaches achieve also the state-of-the-art among frame-recursive methods \cite{fnlk,Arias2019-bnlk}. These methods compute the current frame using only the current noisy frame and the previous denoised frame. They achieve lower results than non-recursive methods, but have a lower memory footprint and (potentially) lower computational cost.

\begin{figure*}
\centering
\includegraphics[width=.9\linewidth]{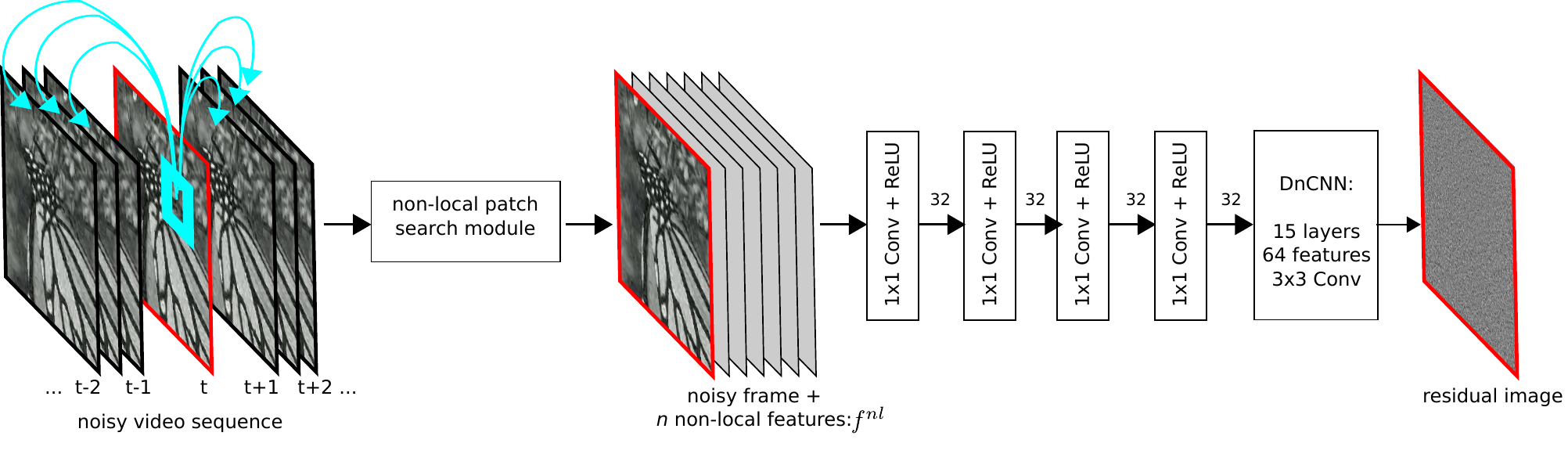}
   \caption{The architecture of the proposed method. The first module performs a patch-wise nearest neighbor search across neighboring frames. Then, the current frame, and the feature vectors $f^{nl}$ of each pixel (the center pixels of the nearest neighbors) are fed into the network. The first four layers of the network perform $1\times1$ convolutions with 32 feature maps. The resulting  feature maps are the input of a simplified  DnCNN~\cite{Zhang2017BeyondDenoising} network with 15 layers.}
\label{fig:architecture}
\end{figure*}

\paragraph{Contributions.}

In this work we propose a non-local architecture for image and video denoising that does not suffer from the restrictions of trainable inference networks.

The method first computes for each image patch the $n$ most similar neighbors in a rectangular spatio-temporal search window and gathers the center pixel of each similar patch forming a feature vector which is assigned to each image location. This results in an image with $n$ channels, which is fed to a CNN trained to predict the clean image from this high dimensional vector. We trained our network for grayscale and color video denoising. Practically training this architecture is made possible by a GPU implementation of the patch search that allows computing the nearest neighbors efficiently.
The self-similarity present temporally in videos enables strong denoising results with our proposal.

To summarize our contributions, in this paper we present a new video denoising CNN method incorporating non-local information in a simple way. To the best of our knowledge, the present work is the first CNN-based video denoising method to attain state-of-the-art results.%

\section{Proposed method}

Let $u$ be a video and $u(x,t)$ denote its value at position $x$ in frame $t$. We observe $v$, a noisy version of $u$ contaminated by additive white Gaussian noise:
\[v = u + r,\]
where $r(x,t)\sim \mathcal N(0,\sigma^2)$.

Our video denoising network processes the video frame by frame. Before it is fed to the network, each frame is processed by a non-local patch search module which computes a non-local feature vector at each image position. A diagram of the proposed network is shown in Figure~\ref{fig:architecture}.

\subsection{Non-local features}

Let $P_{x,t}v$ be a patch centered at pixel $x$ in frame $t$. The patch search module computes the distances between the patch $P_{x,t}v$ and the patches in a 3D rectangular search region $\mathcal R_{x,t}$ centered at $(x,t)$ of size $w_s\times w_s \times w_t$, where $w_s$ and $w_t$ are the spatial and temporal sizes. The positions of these $n$ similar patches are $(x_i, t_i)$ (ordered according to a criterion specified later). Note that $(x_1,t_1) = (x,t)$.

The pixel values at those positions are gathered as an $n$-dimensional non-local feature vector 
\[f^{\text{nl}}(x,t) = [v(x_1, t_1),...,v(x_n,t_n)].\]
The image of non-local features $f^{\text{nl}}$
is considered as a 3D tensor with $n$ channels. This is the input to the network. Note that the first channel of the feature images corresponds to the noisy image $v$.

\subsection{Network architecture}

Our network can be divided in two stages: a non-local stage and a local stage. The non-local stage consists of four $1\times 1$ convolution layers with 32 kernels. The rationale for these layers is to allow the network to compute pixel-wise features out of the raw non-local features $f^{nl}$ at the input.

The second stage receives the features computed by the first stage.  It consists of 14 layers with 64 $3\times 3$ convolution kernels, followed by batch normalization and ReLU activations. The output layer is a $3\times 3$ convolution. Its architecture is similar to the DnCNN network introduced in \cite{Zhang2017BeyondDenoising}, although with 15 layers instead of 17 (as in \cite{17-zhang-ffdnet}). As for DnCNN, the network outputs a residual image, which has to be subtracted to the noisy image to get the denoised one. The training loss is the averaged mean square error between the residual and the noise. For RGB videos, we use the same number of layers, but triple the number of features for each layer.

\section{Training and dataset}

\subsection{Datasets}

For the training and validation sets we used a database of short segments of 16 frames extracted from YouTube videos. Only HD videos with Creative Commons license were used. 
From each video we extracted several segments, separated by at least $10s$.
In total the database consists of 16950 segments extracted from 1068 videos, organized in 64 categories (such as antelope, cars, factory, etc.).
The segments were downscaled to have 540 lines and, when training the grayscale networks, converted to grayscale. An antialising filter was applied before downscaling. To avoid dataset biases, we randomized the filter width. %
We separated 6\% of the videos of the database for the validation (one video for each category).

For training we ignored the first and last frames of each segment for which the 3D patch search window did not fit in the video. For grayscale networks the images were converted to grayscale, before the synthetic Gaussian noise was added. 

During validation we only considered the central frame of each sequence. The resulting validation score is thus computed on 503 sequences (1 frame each).
\footnote{The code to reproduce our results and the database can be found at \url{https://github.com/axeldavy/vnlnet}.}

For testing we used two datasets. One of them is a set of seven sequences from the Derf's Test Media
collection\footnote{\url{https://media.xiph.org/video/derf}} used in~\cite{arias2018comparison}.
This set used exactly the processing pipeline used in~\cite{arias2018comparison}: The original videos are RGB of size $1920 \times 1080$, and sequences of 100 frames were extracted and down-sampled by a factor two (the resolution is thus $960 \times 540$). The grayscale versions were obtained by averaging the channels.
The second dataset is the \texttt{test-dev} split of the DAVIS video segmentation challenge~\cite{davis}. It consists of 30 videos having between 25 and 90 frames. The videos are stored as sequences of JPEG images. There are two versions of the dataset: the full resolution (ranging between HD and 4K) and 480p. We used the full resolution set and applied our own downscaling to 540 rows. In this way we reduced the compression artifacts.

\subsection{Epochs}

At each training epoch a new realization of the noise is added to generate the noisy samples. To speed the training up, we pre-compute the non-local patch search on every video (after noise generation). A random set of (spatio-temporal) patches is drawn from the dataset to generate the mini-batches.
We only consider patches such that the $w_s\times w_s\times w_t$ search window fits in the video (for instance, we exclude the first and last $w_t/2$ frames). At testing time, we simply extended the video by mirroring it at the start and the end of the sequence. An epoch comprised 14000 batches of size 128, composed of image patches of size $44\times44$. We trained for 20 epochs with Adam~\cite{kingma2014adam} and reduced the learning rate at epochs 12 and 17 (from $1e^{-3}$ to $1e^{-4}$ and $1e^{-6}$ respectively). Training a network took 16 hours on an NVIDIA TITAN V for grayscale videos, and 72 hours for color videos.

\section{Experimental results}

We will first show some experiments to highlight relevant aspects of the proposed approach. Then we compare with the state-of-the-art.

\begin{table}[htp!]
	\begin{center}
		{\small
		\renewcommand{\tabcolsep}{1.0mm}
		\renewcommand{\arraystretch}{1.}
		\begin{tabular}{| l | c | c | c |}
			\hline
			Method & No patch & Without oracle & With oracle              \\\hline
            PSNR & 31.24 & 31.28 & 31.85 
			  \\\hline
		\end{tabular}}
	\end{center}
    \caption{PSNR on the CBSD68 dataset (noise standard deviation of 25) for the proposed method on still images. Two variants of our method and a baseline (``No patch'') are compared. ``No patch'' corresponds to the baseline CNN with no nearest neighbor information. The other two versions collect 9 neighbors by comparing $9\times9$ patches. But while the former searches them on the noisy image, the latter determines the patch position on the noise-free image (oracle). In both cases the pixel values for the non-local features are taken from the noisy image.}
    \label{tab:image_tests}
\end{table}

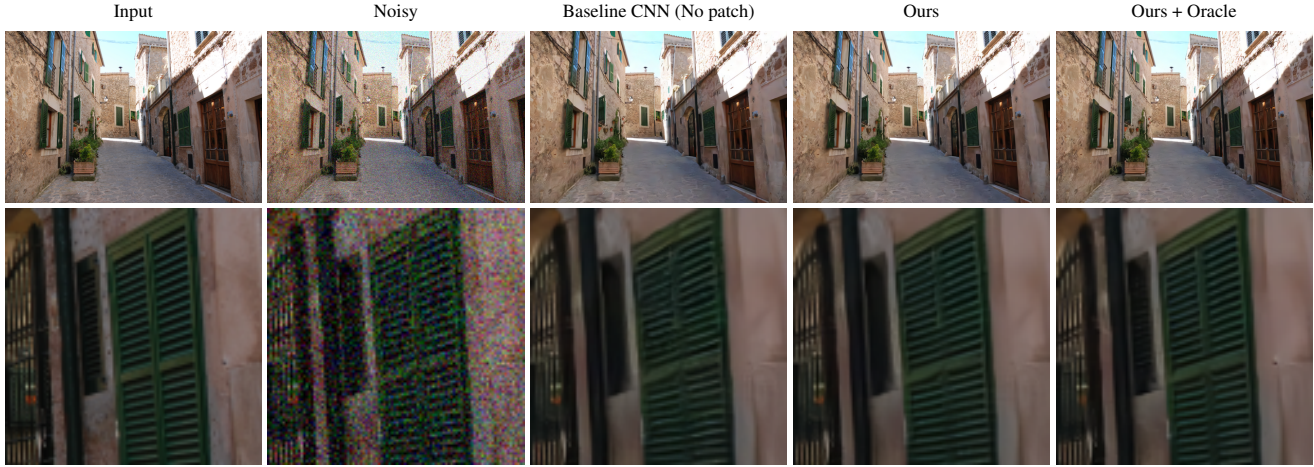
\begin{figure*}
\centering

\begin{tikzpicture}
  \newlength{\nextfigheight}
  \newlength{\figwidth}
  \newlength{\figsep}
  \setlength{\nextfigheight}{0cm}
  \setlength{\figwidth}{0.195\textwidth}
  \setlength{\figsep}{0.2\textwidth}
    \node[anchor=south, inner sep=0] (input) at (0,\nextfigheight) {\includegraphics[width=\figwidth]{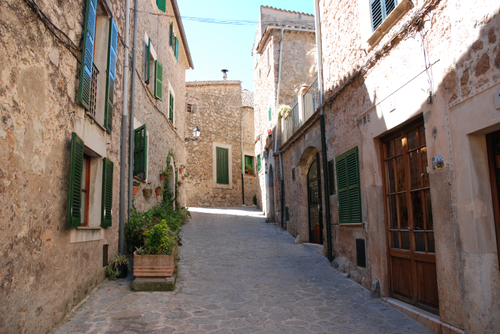}};
\node[anchor=south, inner sep=0] (noisy) at (\figsep,\nextfigheight){\includegraphics[width=\figwidth]{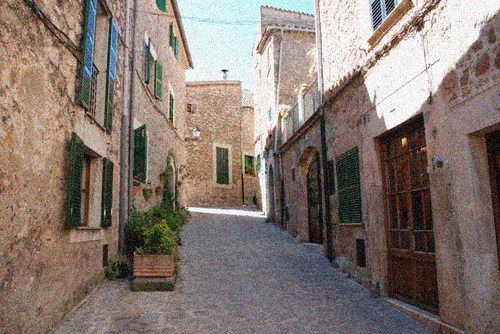}};
\node[anchor=south, inner sep=0] (DNCNN)  at (2*\figsep,\nextfigheight){\includegraphics[width=\figwidth]{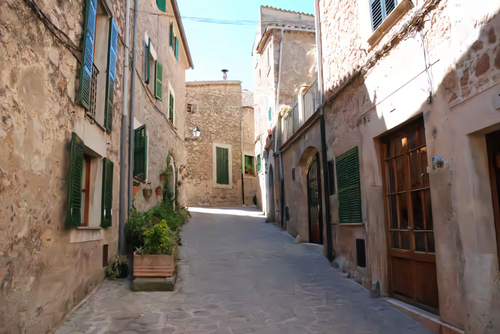}};
\node[anchor=south, inner sep=0] (Ours)  at (3*\figsep,\nextfigheight){\includegraphics[width=\figwidth]{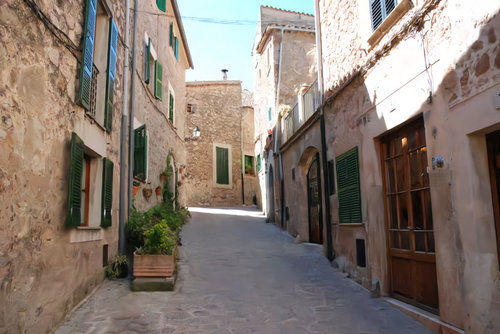}};
\node[anchor=south, inner sep=0] (Oracle)  at (4*\figsep,\nextfigheight){\includegraphics[width=\figwidth]{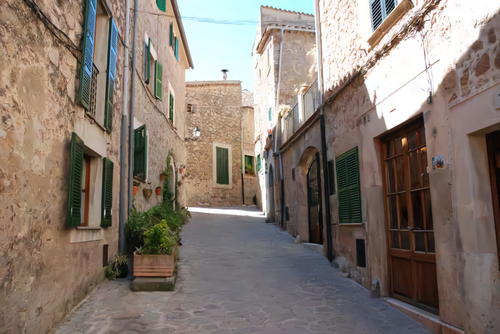}};

    \node [anchor=south] at (input.north) {\scriptsize Input};
    \node [anchor=south] at (noisy.north) {\scriptsize \vphantom{p}Noisy\vphantom{p}};
    \node [anchor=south] at (DNCNN.north) {\scriptsize \vphantom{p}Baseline CNN (No patch)\vphantom{p}};
    \node [anchor=south] at (Ours.north) {\scriptsize \vphantom{p}Ours\vphantom{p}};
    \node [anchor=south] at (Oracle.north) {\scriptsize \vphantom{p}Ours + Oracle\vphantom{p}};

\addtolength{\nextfigheight}{-\figsep}

\node[anchor=south, inner sep=0] (input_crop1) at (0,\nextfigheight) {\includegraphics[width=\figwidth]{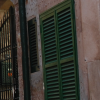}};
\node[anchor=south, inner sep=0] (noisy_crop1) at (\figsep,\nextfigheight){\includegraphics[width=\figwidth]{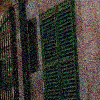}};
\node[anchor=south, inner sep=0] (DNCNN_crop1)  at (2*\figsep,\nextfigheight){\includegraphics[width=\figwidth]{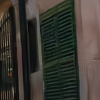}};
\node[anchor=south, inner sep=0] (Ours_crop1)  at (3*\figsep,\nextfigheight){\includegraphics[width=\figwidth]{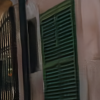}};
\node[anchor=south, inner sep=0] (Oracle_crop1)  at (4*\figsep,\nextfigheight){\includegraphics[width=\figwidth]{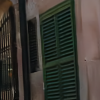}};

 \end{tikzpicture}

\caption{Results on a color image (noise standard deviation of 25). The compared methods are the ones introduced in Table~\ref{tab:image_tests}.}
\label{fig:image_result}
\end{figure*}

\paragraph{The untapped potential of non-locality.} Although the focus of this work is in video denoising, it is still interesting to study the performance of the proposed non-local CNN on images. Figure~\ref{fig:image_result} shows a comparison of a baseline CNN (a 15 layer version of DnCNN \cite{Zhang2017BeyondDenoising}, as in our network) and a version of our method trained for still image denoising (it collects $9$ neighbors by comparing $9\times9$ patches).
The results  with and without non-local information are very similar, this is confirmed on Table~\ref{tab:image_tests}. The only difference is visible on very self-similar parts like the blinds that are shown in the detail of Figure~\ref{fig:image_result}. 
The average PSNR on the CBSD68 dataset \cite{MartinFTM01,Zhang2017BeyondDenoising} (noise with $\sigma=25$) obtained for the baseline CNN is of 31.24dB. The non-local CNN only leads to a 0.04dB improvement (31.28dB).
The figure and table also show the result of an oracular method: the nearest neighbor search is performed on the noise-free image, though the pixel values are taken from the noisy image. 
The oracular results show that non-locality has a great potential to improve the results of CNNs. The oracular method obtains an average PSNR of 31.85dB, 0.6dB over the baseline.
However, this improvement is hindered by the difficulty of finding accurate matches in the presence of noise. 
A way to reduce the matching errors is to use larger patches. But on images, larger patches have fewer similar patches. 
In  contrast, as we will see below, the temporal redundancy of videos allows using very large patches.

\subsection{Parameter tuning}

Non-local search has three main parameters: The patch size, the number of retained matches and the number of frames in the search region. We expect the best matches to be past or future versions of the current patch, so we set the number of matches as the number of frames on which we search.

\begin{figure*}
\centering
\begin{tikzpicture}
  \setlength{\nextfigheight}{0cm}
  \setlength{\figwidth}{0.12\textwidth}
  \setlength{\figsep}{0.125\textwidth}
    \node[anchor=south, inner sep=0] (input) at (0,\nextfigheight) {\includegraphics[width=\figwidth]{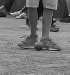}};
\node[anchor=south, inner sep=0] (noisy) at (\figsep,\nextfigheight){\includegraphics[width=\figwidth]{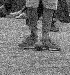}};
\node[anchor=south, inner sep=0] (n0)  at (2*\figsep,\nextfigheight){\includegraphics[width=\figwidth]{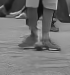}};
\node[anchor=south, inner sep=0] (n1)  at (3*\figsep,\nextfigheight){\includegraphics[width=\figwidth]{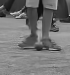}};
\node[anchor=south, inner sep=0] (n2)  at (4*\figsep,\nextfigheight){\includegraphics[width=\figwidth]{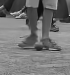}};
\node[anchor=south, inner sep=0] (n3)  at (5*\figsep,\nextfigheight){\includegraphics[width=\figwidth]{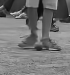}};
\node[anchor=south, inner sep=0] (n4)  at (6*\figsep,\nextfigheight){\includegraphics[width=\figwidth]{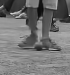}};
\node[anchor=south, inner sep=0] (n5)  at (7*\figsep,\nextfigheight){\includegraphics[width=\figwidth]{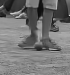}};

    \node [anchor=south] at (input.north) {\scriptsize Input};
    \node [anchor=south] at (noisy.north) {\scriptsize \vphantom{p}Noisy\vphantom{p}};
    \node [anchor=south] at (n0.north) {\scriptsize \vphantom{p}No patch\vphantom{p}};
    \node [anchor=south] at (n1.north) {\scriptsize \vphantom{p}Patch width 9\vphantom{p}};
    \node [anchor=south] at (n2.north) {\scriptsize \vphantom{p}Patch width 15\vphantom{p}};
    \node [anchor=south] at (n3.north) {\scriptsize \vphantom{p}Patch width 21\vphantom{p}};
    \node [anchor=south] at (n4.north) {\scriptsize \vphantom{p}Patch width 31\vphantom{p}};
    \node [anchor=south] at (n5.north) {\scriptsize \vphantom{p}Patch width 41\vphantom{p}};

\addtolength{\nextfigheight}{-\figsep}

\node[anchor=south, inner sep=0] () at (0,\nextfigheight) {\includegraphics[width=\figwidth]{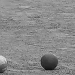}};
\node[anchor=south, inner sep=0] () at (\figsep,\nextfigheight){\includegraphics[width=\figwidth]{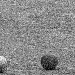}};
\node[anchor=south, inner sep=0] ()  at (2*\figsep,\nextfigheight){\includegraphics[width=\figwidth]{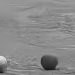}};
\node[anchor=south, inner sep=0] ()  at (3*\figsep,\nextfigheight){\includegraphics[width=\figwidth]{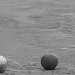}};
\node[anchor=south, inner sep=0] ()  at (4*\figsep,\nextfigheight){\includegraphics[width=\figwidth]{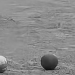}};
\node[anchor=south, inner sep=0] ()  at (5*\figsep,\nextfigheight){\includegraphics[width=\figwidth]{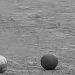}};
\node[anchor=south, inner sep=0] ()  at (6*\figsep,\nextfigheight){\includegraphics[width=\figwidth]{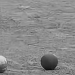}};
\node[anchor=south, inner sep=0] ()  at (7*\figsep,\nextfigheight){\includegraphics[width=\figwidth]{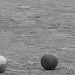}};

 \end{tikzpicture}
\caption{Example of denoised results with our method when changing the patch size, respectively no patch search, $9\times9$, $15\times15$, $21\times21$, $31\times31$ and $41\times41$ patches. The 3D search window has 15 frames for these experiments.}
\label{fig:figB}
\end{figure*}
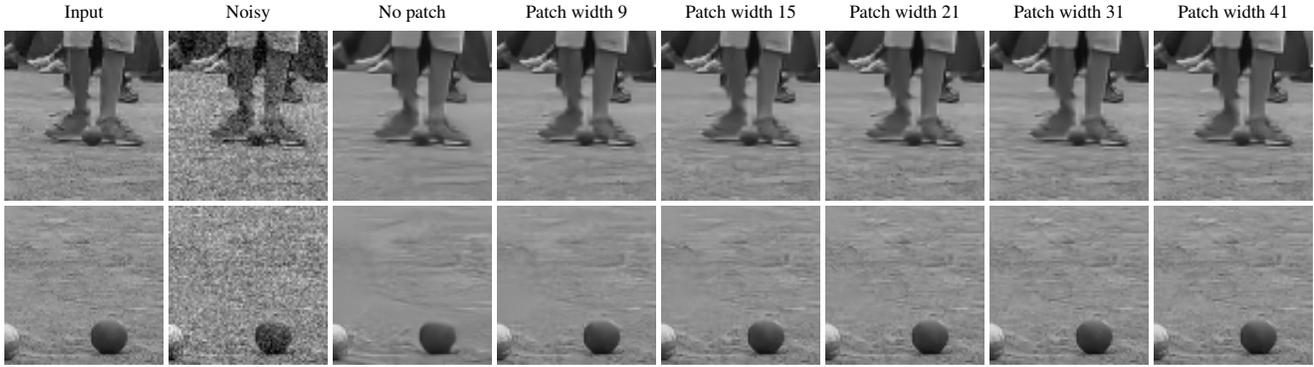

\begin{table}[htp!]
	\begin{center}
		{\small
		\renewcommand{\tabcolsep}{1.0mm}
		\renewcommand{\arraystretch}{1.}
		\begin{tabularx}{\linewidth}{| X | c | c | c | c | c | c |}
			\hline
			Patch size & no patch & $9\!\times\!9$ & $15\!\times\!15$ & $21\!\times\!21$ & $31\!\times\!31$ & $41\!\times\!41$              \\\hline
            PSNR & 33.75 & 35.62 & 36.40 & 36.84 & 37.11 & 37.22
			  \\\hline
		\end{tabularx}}
	\end{center}
    \caption{Impact of the patch size on the PSNR computed on the validation set (noise standard deviation of 20). The tested sizes are $9\times9$, $15\times15$, $21\times21$, $31\times31$ and $41\times41$. No patch corresponds to the baseline simplified DnCNN.}
    \label{tab:patch_width_impact_graph}
\end{table}

\begin{figure*}
\centering

\begin{tikzpicture}
  \setlength{\nextfigheight}{0cm}
  \setlength{\figwidth}{0.137\textwidth}
  \setlength{\figsep}{0.142\textwidth}
    \node[anchor=south, inner sep=0] (input) at (0,\nextfigheight) {\includegraphics[width=\figwidth]{figBC/input_crop1.png}};
\node[anchor=south, inner sep=0] (noisy) at (\figsep,\nextfigheight){\includegraphics[width=\figwidth]{figBC/noisy_crop1.png}};
\node[anchor=south, inner sep=0] (n0)  at (2*\figsep,\nextfigheight){\includegraphics[width=\figwidth]{figBC/noise20_mainline_crop1.png}};
\node[anchor=south, inner sep=0] (n1)  at (3*\figsep,\nextfigheight){\includegraphics[width=\figwidth]{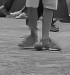}};
\node[anchor=south, inner sep=0] (n2)  at (4*\figsep,\nextfigheight){\includegraphics[width=\figwidth]{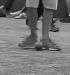}};
\node[anchor=south, inner sep=0] (n3)  at (5*\figsep,\nextfigheight){\includegraphics[width=\figwidth]{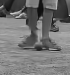}};
\node[anchor=south, inner sep=0] (n4)  at (6*\figsep,\nextfigheight){\includegraphics[width=\figwidth]{figBC/noise20_nn15_p41w_crop1.png}};

    \node [anchor=south] at (input.north) {\scriptsize Input};
    \node [anchor=south] at (noisy.north) {\scriptsize \vphantom{p}Noisy\vphantom{p}};
    \node [anchor=south] at (n0.north) {\scriptsize \vphantom{p}No Patch\vphantom{p}};
    \node [anchor=south] at (n1.north) {\scriptsize \vphantom{p}3 Neighbors\vphantom{p}};
    \node [anchor=south] at (n2.north) {\scriptsize \vphantom{p}7 Neighbors\vphantom{p}};
    \node [anchor=south] at (n3.north) {\scriptsize \vphantom{p}11 Neighbors\vphantom{p}};
    \node [anchor=south] at (n4.north) {\scriptsize \vphantom{p}15 Neighbors\vphantom{p}};

\addtolength{\nextfigheight}{-\figsep}

\node[anchor=south, inner sep=0] () at (0,\nextfigheight) {\includegraphics[width=\figwidth]{figBC/input_crop2.png}};
\node[anchor=south, inner sep=0] () at (\figsep,\nextfigheight){\includegraphics[width=\figwidth]{figBC/noisy_crop2.png}};
\node[anchor=south, inner sep=0] ()  at (2*\figsep,\nextfigheight){\includegraphics[width=\figwidth]{figBC/noise20_mainline_crop2.png}};
\node[anchor=south, inner sep=0] ()  at (3*\figsep,\nextfigheight){\includegraphics[width=\figwidth]{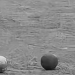}};
\node[anchor=south, inner sep=0] ()  at (4*\figsep,\nextfigheight){\includegraphics[width=\figwidth]{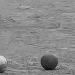}};
\node[anchor=south, inner sep=0] ()  at (5*\figsep,\nextfigheight){\includegraphics[width=\figwidth]{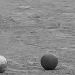}};
\node[anchor=south, inner sep=0] ()  at (6*\figsep,\nextfigheight){\includegraphics[width=\figwidth]{figBC/noise20_nn15_p41w_crop2.png}};

 \end{tikzpicture}
\caption{Example of denoised results with our method when changing the number of frames considered in the 3D search window (respectively no patch search, 3, 7, 11 and 15). $41\times41$ patches were used for these experiments.}
\label{fig:figC}
\end{figure*}

\begin{table}[htp!]
	\begin{center}
		{\small
		\renewcommand{\tabcolsep}{1.0mm}
		\renewcommand{\arraystretch}{1.}
		\begin{tabularx}{\linewidth}{| X | c | c | c | c | c |}
			\hline
			\# search frames & \hspace{1em} no patch \hspace{1em} & 3 & 7 & 11 & 15              \\\hline
            PSNR & 33.75 & 35.35 & 36.50 & 36.97 & 37.22
			  \\\hline

		\end{tabularx}}
	\end{center}
    \caption{Impact of the number of frames considered in the 3D search window, on the PSNR computed on the validation set for a noise standard deviation of 20. (respectively no patch search, 3, 7, 11 and 15)}
    \label{tab:numneighbors_impact_graph}
\end{table}

\begin{figure}
\centering
\begin{tikzpicture}
  \setlength{\nextfigheight}{0cm}
  \setlength{\figwidth}{0.24\columnwidth}
  \setlength{\figsep}{0.25\columnwidth}
    \node[anchor=south, inner sep=0] (input) at (0,\nextfigheight) {\includegraphics[width=\figwidth]{figBC/input_crop1.png}};
\node[anchor=south, inner sep=0] (noisy) at (\figsep,\nextfigheight){\includegraphics[width=\figwidth]{figBC/noisy_crop1.png}};
\node[anchor=south, inner sep=0] (n0)  at (2*\figsep,\nextfigheight){\includegraphics[width=\figwidth]{figBC/noise20_nn15_p41w_crop1.png}};
\node[anchor=south, inner sep=0] (n1)  at (3*\figsep,\nextfigheight){\includegraphics[width=\figwidth]{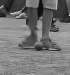}}; 

    \node [anchor=south] at (input.north) {\scriptsize Input};
    \node [anchor=south] at (noisy.north) {\scriptsize \vphantom{p}Noisy\vphantom{p}};
    \node [anchor=south] at (n0.north) {\scriptsize \vphantom{p}No restriction\vphantom{p}};
    \node [anchor=south] at (n1.north) {\scriptsize \vphantom{p}One patch per frame \vphantom{p}};

\addtolength{\nextfigheight}{-\figsep}

\node[anchor=south, inner sep=0] () at (0,\nextfigheight) {\includegraphics[width=\figwidth]{figBC/input_crop2.png}};
\node[anchor=south, inner sep=0] () at (\figsep,\nextfigheight){\includegraphics[width=\figwidth]{figBC/noisy_crop2.png}};
\node[anchor=south, inner sep=0] ()  at (2*\figsep,\nextfigheight){\includegraphics[width=\figwidth]{figBC/noise20_nn15_p41w_crop2.png}};
\node[anchor=south, inner sep=0] ()  at (3*\figsep,\nextfigheight){\includegraphics[width=\figwidth]{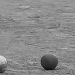}};

 \end{tikzpicture}
\caption{Example of denoised results with our method when allowing patches to be selected anywhere on the 3D search region, or when having exactly one patch neighbor per frame. $41\times41$ patches and a search region of 15 frames were used for these experiments.}
\label{fig:restriction}
\end{figure}

\begin{table}[htp!]
	\begin{center}
		{\small
		\renewcommand{\tabcolsep}{1.0mm}
		\renewcommand{\arraystretch}{1.}
		\begin{tabularx}{\linewidth}{| X | c | c |}
			\hline
			Patch search & \hspace{1em} no restriction \hspace{1em} & one neighbor per frame              \\\hline
            PSNR & 37.22 & 37.46
			  \\\hline

		\end{tabularx}}
	\end{center}
    \caption{Impact of allowing patches to be selected anywhere on the 3D search region, or having exactly one neighbor per frame. The PSNR is computed on the validation set (noise standard deviation of 20), with a patch size $41\times41$ and a search region of 15 frames.}
    \label{tab:restriction}
\end{table}

In Table~\ref{tab:patch_width_impact_graph}, we explore the impact of the patch size used for the matching. Figure~\ref{fig:figB} shows visual results corresponding to each parameter. Surprisingly, we obtain better and better results by increasing the size of the patches. The main reason for this is that the match precision is improved, as the impact of 
noise on the patch distance shrinks. The bottom row of Figure~\ref{fig:figB} shows an area of the ground only affected by slight camera motion and on the top row an area with complex motion (a person moving his feet). We can see that the former is clearly better denoised using large patches, while the latter remains unaffected around the motion area. This indicates 
that the network is able to determine when the provided non-local information is not accurate and to fall back to a result similar to DnCNN in this case (single image denoising), which can be noticed on the last row of Figure~\ref{fig:figA}. Further increasing the patch size would result in more areas being processed as single images. As a result, we see that the performance gain from $31\times31$ to 
$41\times41$ is rather small. With such large patches, only matches of the same objects from different frames are likely to be taken as neighbors. Thus we go a step further by enforcing matches to come from different frames, which improves slightly the performance. This is shown on Figure~\ref{fig:restriction} and Table~\ref{tab:restriction}. Note the network is retrained as the patch distribution is impacted. Indeed when no restriction are imposed, the neighbors are sorted by increasing distance. While, in this variant, neighbors are sorted by frame index.

In Table~\ref{tab:numneighbors_impact_graph} and Figure~\ref{fig:figC}, we see the impact of the number of frames used in the search window (and thus the number of nearest neighbors). One can see that the more frames, the better. Increasing the number of frames beyond 15 (7 past, current, and 7 future) does not justify the small increase of performance. Foreground moving objects are unlikely to get good neighbors for the selected patch size, unlike background objects, thus it 
comes to no surprise that the visual quality of the background improves with the number of patches, while foreground moving objects (for example the legs on Figure~\ref{fig:figC}) do not improve much.

In the following experiments, we shall use $41\times41$ patches and 15 frames. Another parameter for non-local search is the spatial width of the search window, which we set to 41 pixels (the center pixel of the tested patches must reside inside this region). We trained grayscale and color networks for AGWN of $\sigma$ 10, 20 and 40. To highlight the fact that a CNN method can adapt to many noise types, unlike traditional 
methods, we also trained a grayscale network for Gaussian noise correlated by a $3\times3$ box kernel such that the final standard deviation is $\sigma=20$, and 25\% uniform Salt and Pepper noise (removed pixels are replaced by random uniform noise).

\subsection{Comparison with state-of-the-art}

\begin{table*}[htp!]
	\begin{center}
		{\small
		\renewcommand{\tabcolsep}{1.0mm}
		\renewcommand{\arraystretch}{1.}
		\begin{tabular}{| c | l |c | c | c | c | c | c | c | c |}
			\hline
			\rule{0pt}{6pt}$\sigma$
			& Method               &    {crowd}          &    {park joy}        & {pedestrians}        &    {station}        &    {sunflower}       &     {touchdown}     &  {tractor}          & average              \\\hline
			\multirow{1}{*}{$10$}
			& SPTWO                &    36.57  /    .9651  &    35.87  /    .9570  &    41.02  /    .9725  &    41.24  /    .9697  &    42.84  /    .9824  &    40.45  /    .9557  &    \SB{38.92}  /    .9701  &    39.56  /    .9675  \\
			& VBM3D            &    35.76  /    .9589  &    35.00  /    .9469  &    40.90  /    .9674  &    39.14  /    .9651  &    40.13  /    .9770  &    39.25  /    .9466  &    37.51  /    .9575  &    38.24  /    .9599  \\
			& VBM4D            &    36.05  /    .9535  &    35.31  /    .9354  &    40.61  /    .9712  &    40.85  /    .9466  &    41.88  /    .9696  &    39.79  /    .9440  &    37.73  /    .9533  &    38.88  /    .9534  \\
			& VNLB                 & \B{37.24} / \SB{.9702} & \B{36.48} / \SB{.9622} & \B{42.23} / \B{.9782} & \SB{42.14} / \B{.9771} & \SB{43.70} / \SB{.9850} & \B{41.23} / \B{.9615} & \B{40.20} / \B{.9773} & \B{40.57} / \B{.9731} \\
			& DnCNN                &    34.39  /    .9455  &    33.82  /    .9329  &    39.46  /    .9641  &    37.89  /    .9412  &    40.20  /    .9702  &    38.28  /    .9269  &    36.91  /    .9568  &    37.28  /    .9482  \\
			& VNLnet                &    \SB{37.00}  /    \B{.9727}  &    \SB{36.39}  /    \B{.9665}  &    \SB{41.96}  /    \SB{.9779}  &    \B{42.44}  /    \SB{.9766}  &    \B{43.76}  /    \B{.9861}  &    \SB{41.05}  /    \SB{.9609}  &    38.89  /    \SB{.9718}  &    \SB{40.21}  /    \SB{.9732}  \\\hline
			\multirow{1}{*}{$20$} 
			& SPTWO                &    32.94  /    .9319  &    32.35  / .9161 &    37.01  /    .9391  &    38.09  /    \SB{.9461}  &    38.83  /    .9593  & \B{37.55} / \B{.9287} &    35.15  /    .9363  &    35.99  /    .9368  \\
			& VBM3D            &    32.34  /    .9093  &    31.50  /    .8731  &    37.06  /    .9423  &    35.91  /    .9007  &    36.25  /    .9393  &    36.17  /    .9065  &    33.53  /    .8991  &    34.68  /    .9100  \\
			& VBM4D            &    32.40  /    .9126  &    31.60  /    .8832  &    36.72  /    .9344  &    36.84  /    .9224  &    37.78  /    .9517  &    36.44  /    .9034  &    33.95  /    .9104  &    35.10  /    .9169  \\
			& VNLB                 & \B{33.49} / \SB{.9335} & \SB{32.80} /    \SB{.9154}  & \B{38.61} / \B{.9583} & \B{38.78} / \B{.9470} & \SB{39.82} / \SB{.9698} &   \SB{37.47} /    \SB{.9220}  & \B{36.67} / \B{.9536} & \B{36.81} / \B{.9428} \\
			& DnCNN                &    30.47  /    .8890  &    30.03  /    .8625  &    35.81  /    .9302  &    34.37  /    .8832  &    36.19  /    .9361  &    35.35  /    .8782  &    32.99  /    .9019  &    33.60  /    .8973  \\
			& VNLnet                &    \SB{33.40}  /  \B{.9415}  &    \B{32.84}  /    \B{.9271} &    \SB{38.32}  /    \SB{.9565}  &    \SB{38.49}  /    .9454  &    \B{39.88}  /    \B{.9700}  &    37.11  /    .9102  &    \SB{35.23}  /    \SB{.9390}  &    \SB{36.47}  /    \SB{.9414}  \\\hline
			\multirow{1}{*}{$40$} 
			& SPTWO                &    29.02  /    .8095  &    28.79  /    .8022  &    31.32  /    .7705  &    32.37  /    .7922  &    32.61  /    .7974  &    31.80  /    .7364  &    30.61  /    .8223  &    30.93  /    .7901  \\
			& VBM3D            &    28.73  /    .8295  &    27.93  /    .7663  &    33.00  /    .8828  &    32.57  /    .8239  &    32.39  /    .8831  &    33.38  /    \SB{.8624}  &    29.80  /    .8039  &    31.11  /    .8360  \\
			& VBM4D            &    28.72  /    .8339  &    27.99  /    .7751  &    32.62  /    .8683  &    32.93  /    .8441  &    33.66  /    .8999  &    33.68  /    .8603  &    30.20  /    .8205  &    31.40  /    .8432  \\
			& VNLB                 & \B{29.88} / \SB{.8682} & \SB{29.28} / \SB{.8309} & \B{34.68} /    \B{.9167}  & \B{34.65} / \B{.8871} &    \B{35.44}  /    \B{.9329}  &    \B{34.18}  /    \B{.8712}  & \B{32.58} / \B{.8921} & \B{32.95} / \B{.8856} \\
			& DnCNN                &    26.85  /    .7979  &    26.65  /    .7525  &    32.01  /    .8660  &    30.96  /    .7899  &    32.13  /    .8705  &    32.78  /    .8346  &    29.25  /    .7976  &    30.09  /    .8156  \\
			& VNLnet                &    \SB{29.69}  /    \B{.8727}  &    \B{29.29}  /    \B{.8397}  &    \SB{34.21}  /    \SB{.9089}  &    \SB{33.96}  /    \SB{.8686}  &    \SB{35.12}  /    \SB{.9224}  &    \SB{33.88}  /    .8495  &    \SB{31.41}  /    \SB{.8647}  &    \SB{32.51}  /    \SB{.8752}  \\\hline
		\end{tabular}}
	\end{center}
    \caption{Quantitative denoising results (PSNR and SSIM) for seven grayscale test
		sequences of size $960\times 540$ from the \textit{Derf's Test Media
collection} on several state-of-the-art video denoising algorithms versus DnCNN and our method. Three noise standard deviations $\sigma$ are tested (10, 20 and 40). Compared methods are SPTWO~\cite{buades2016patch}, VBM3D~\cite{Dabov2007v}, VBM4D~\cite{Maggioni2011}, VNLB~\cite{Arias2015}, DnCNN~\cite{Zhang2017BeyondDenoising} and VNLnet (ours). We highlighted the best performance in black and the second best in brown.}
	\label{tab:psnr-classic-gray}
\end{table*}

\begin{table*}[htp!]
	\begin{center}
		{\small
		\renewcommand{\tabcolsep}{1.0mm}
		\renewcommand{\arraystretch}{1.}
		\begin{tabular}{| c | l |c | c | c | c | c | c | c | c |}
			\hline
			\rule{0pt}{6pt}$\sigma$
			& Method               &    {crowd}          &    {park joy}        & {pedestrians}        &    {station}        &    {sunflower}       &     {touchdown}     &  {tractor}          & average              \\\hline
			\multirow{1}{*}{$10$}
            & VBM3D & 36.03 / .9625 & 35.01 / .9451  & 41.19 / .9738  & 38.53 / .9463  & 39.58 / .9599 &   39.91 / .9486 & 37.10 / .9555  & 38.19 / .9560 \\
			& VNLB & \B{38.33} / \B{.9773}  & \B{37.09} / \B{.9708}  &  \B{42.77} / \SB{.9800} & \B{42.83} / \B{.9784}  & \B{43.23} / \B{.9820} &  \SB{42.16} / \SB{.9677}  & \B{40.07} / \B{.9760}  & \B{40.93} / \B{.9760} \\
			& DnCNN & 35.41 / .9576 &  34.37 / .9454 & 40.26 / .9701  & 38.73 / .9536  & 40.10 / .9675 &   39.77 / .9485 & 37.37 / .9600  & 38.00 / .9575 \\
			& VNLnet & \SB{37.74} / \SB{.9758} & \SB{36.63} / \SB{.9690}  & \SB{42.56} / \B{.9805}  & \SB{42.22} / \SB{.9765} & \SB{43.12} / \SB{.9819} &    \B{42.26} / \B{.9705} & \SB{38.90} / \SB{.9699}  & \SB{40.49} / \SB{.9749} \\\hline
			\multirow{1}{*}{$20$} 
            & VBM3D & 32.54 / .9284 & 31.58 / .8930  & 37.73 / .9505  & 35.28 / .8962  & 36.01 / .9250 &  36.89 / .9091  & 33.56 / .9131  &  34.80 / .9165 \\
			& VNLB & \B{34.78} / \SB{.9529} & \B{33.53} / \SB{.9365}  & \B{39.63} / \SB{.9640}  & \B{39.67} / \SB{.9565}  & \SB{39.84} / \SB{.9667} &  \SB{38.80} / \SB{.9353}  & \B{37.08} / \B{.9575}  & \B{37.62} / \SB{.9528} \\
			& DnCNN & 31.49 / .9144 & 30.62 / .8875  & 36.92 / .9459  & 35.44 / .9094  & 36.24 / .9373 &    36.70 / .9064 &  33.65 / .9190 & 34.44 / .9171 \\
			& VNLnet & \SB{34.45} / \B{.9556} & \SB{33.40} / \B{.9407}  & \SB{39.63} / \B{.9670}  & \SB{39.57} / \B{.9592}  & \B{40.06} / \B{.9695} & \B{39.30} / \B{.9465}  &  \SB{35.78} / \SB{.9456} & \SB{37.46} / \B{.9549} \\\hline
			\multirow{1}{*}{$40$} 
			& VBM3D & 29.23 / .8688 & 28.43 / .8152  & 34.11 / .9056 & 32.45 / .8255 & 32.82 / .8792 &    34.17 / .8604 & 30.30 / .8431 & 31.65 / .8568 \\
			& VNLB & \B{31.24} / \SB{.9052} & \SB{30.23} / \SB{.8730}  & \SB{35.97} / \SB{.9291} & \SB{35.88} / \SB{.9074}  & \SB{35.77} / \SB{.9285} &  \SB{35.19} / \SB{.8719}  & \B{33.47} / \B{.9140}  & \SB{33.97} / \SB{.9042} \\
			& DnCNN & 27.99 / .8429 & 27.50 / .7998 & 33.50 / .9001  & 32.29 / .8350  & 32.58 / .8845  &   33.94 / .8566  & 30.17 / .8454  & 31.14 / .8520 \\
			& VNLnet & \SB{31.13} / \B{.9144} & \B{30.23} / \B{.8848}  & \B{36.19} / \B{.9390}  &  \B{36.12} / \B{.9172} & \B{36.36} / \B{.9411} &  \B{35.64} / \B{.8872}  & \SB{32.44} / \SB{.8985}  & \B{34.02} / \B{.9117} \\\hline
		\end{tabular}}
	\end{center}
    \caption{Quantitative denoising results (PSNR and SSIM) for seven color test
		sequences of size $960\times 540$ from the \textit{Derf's Test Media
collection} on several state-of-the-art video denoising algorithms versus DnCNN and our method. Three noise standard deviations $\sigma$ are tested (10, 20 and 40). Compared methods are VBM3D~\cite{Dabov2007v}, VNLB~\cite{Arias2015}, DnCNN~\cite{Zhang2017BeyondDenoising} and VNLnet (ours). We highlighted the best performance in black and the second best in brown.}
	\label{tab:psnr-classic-color}
\end{table*}

\begin{figure*}
\centering
\begin{tikzpicture}
  \setlength{\nextfigheight}{0cm}
  \setlength{\figwidth}{0.16\textwidth}
  \setlength{\figsep}{0.166\textwidth}
    \node[anchor=south, inner sep=0] (input) at (0,\nextfigheight) {\includegraphics[width=\figwidth]{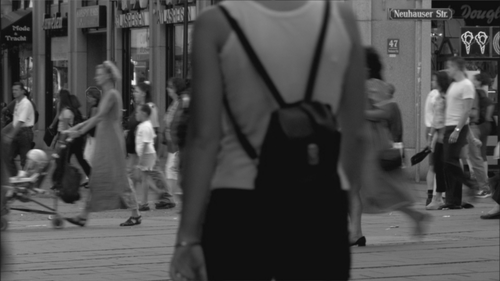}};
\node[anchor=south, inner sep=0] (noisy) at (\figsep,\nextfigheight){\includegraphics[width=\figwidth]{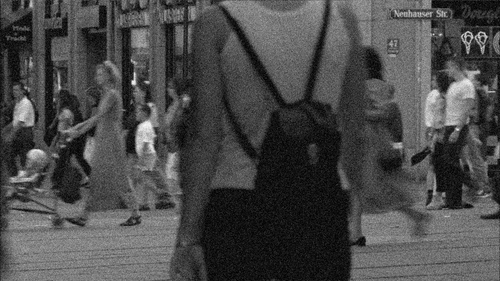}};
\node[anchor=south, inner sep=0] (NLPM)  at (2*\figsep,\nextfigheight){\includegraphics[width=\figwidth]{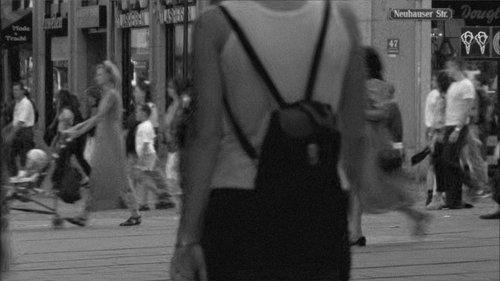}};
\node[anchor=south, inner sep=0] (DNCNN)  at (3*\figsep,\nextfigheight){\includegraphics[width=\figwidth]{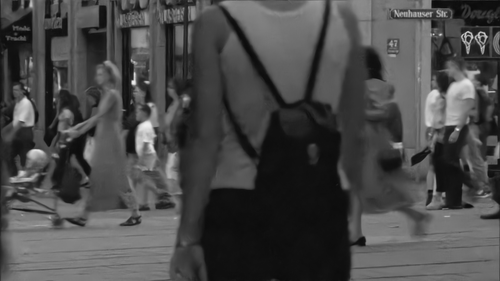}};
\node[anchor=south, inner sep=0] (VNLB)  at (4*\figsep,\nextfigheight){\includegraphics[width=\figwidth]{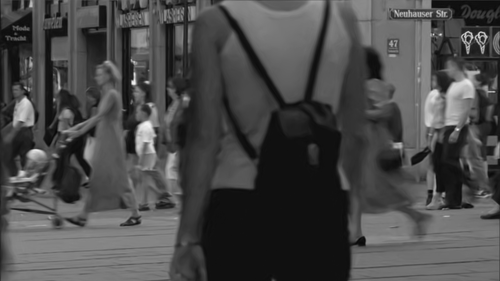}};
\node[anchor=south, inner sep=0] (Ours)  at (5*\figsep,\nextfigheight){\includegraphics[width=\figwidth]{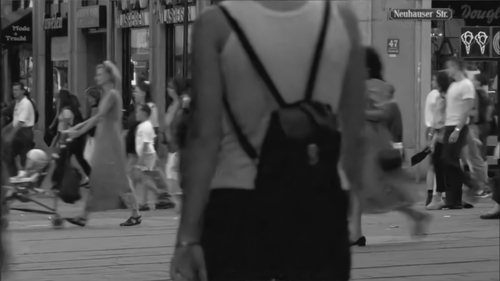}};

    \node [anchor=south] at (input.north) {\scriptsize Input};
    \node [anchor=south] at (noisy.north) {\scriptsize \vphantom{p}Noisy\vphantom{p}};
    \node [anchor=south] at (NLPM.north) {\scriptsize \vphantom{p}\textit{Non-Local Pixel Mean}\vphantom{p}};
    \node [anchor=south] at (DNCNN.north) {\scriptsize \vphantom{p}DnCNN\vphantom{p}};
    \node [anchor=south] at (VNLB.north) {\scriptsize \vphantom{p}VNLB\vphantom{p}};
    \node [anchor=south] at (Ours.north) {\scriptsize \vphantom{p}VNLnet (Ours)\vphantom{p}};

\addtolength{\nextfigheight}{-\figsep}

\node[anchor=south, inner sep=0] (input_crop1) at (0,\nextfigheight) {\includegraphics[width=\figwidth]{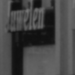}};
\node[anchor=south, inner sep=0] (noisy_crop1) at (\figsep,\nextfigheight){\includegraphics[width=\figwidth]{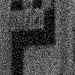}};
\node[anchor=south, inner sep=0] (NLPM_crop1)  at (2*\figsep,\nextfigheight){\includegraphics[width=\figwidth]{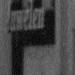}};
\node[anchor=south, inner sep=0] (DNCNN_crop1)  at (3*\figsep,\nextfigheight){\includegraphics[width=\figwidth]{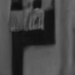}};
\node[anchor=south, inner sep=0] ()  at (4*\figsep,\nextfigheight){\includegraphics[width=\figwidth]{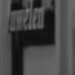}};
\node[anchor=south, inner sep=0] ()  at (5*\figsep,\nextfigheight){\includegraphics[width=\figwidth]{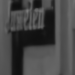}};

\addtolength{\nextfigheight}{-\figsep}

\node[anchor=south, inner sep=0] (input_crop2) at (0,\nextfigheight) {\includegraphics[width=\figwidth]{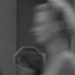}};
\node[anchor=south, inner sep=0] (noisy_crop2) at (\figsep,\nextfigheight){\includegraphics[width=\figwidth]{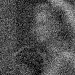}};
\node[anchor=south, inner sep=0] (NLPM_crop2)  at (2*\figsep,\nextfigheight){\includegraphics[width=\figwidth]{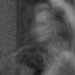}};
\node[anchor=south, inner sep=0] (DNCNN_crop2)  at (3*\figsep,\nextfigheight){\includegraphics[width=\figwidth]{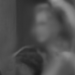}};
\node[anchor=south, inner sep=0] ()  at (4*\figsep,\nextfigheight){\includegraphics[width=\figwidth]{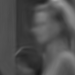}};
\node[anchor=south, inner sep=0] ()  at (5*\figsep,\nextfigheight){\includegraphics[width=\figwidth]{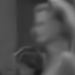}};

 \end{tikzpicture}
 
\caption{Example of denoised result for several algorithms (noise standard deviation of 20). The two crops highlight the results on a non-moving and a moving part of the video. \textit{Non-Local Pixel Mean} corresponds to the average of the output of the non-local search layer.}
\label{fig:figA}
\end{figure*}
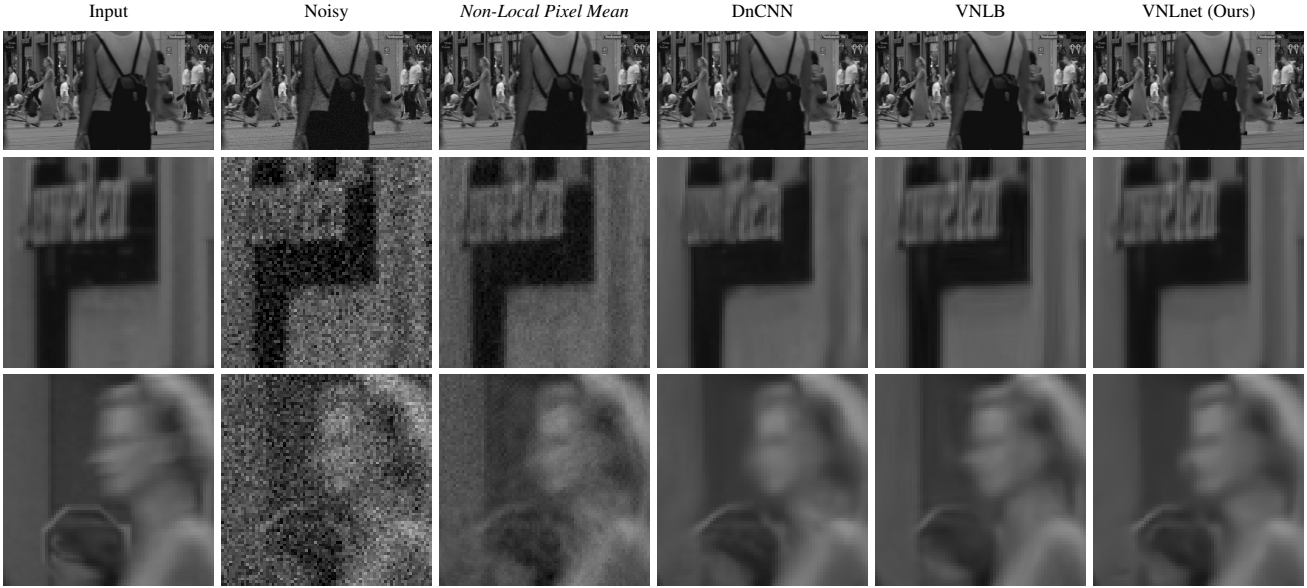

\begin{figure*}
\centering
\begin{tikzpicture}
  \setlength{\nextfigheight}{0cm}
  \setlength{\figwidth}{0.16\textwidth}
  \setlength{\figsep}{0.166\textwidth}
    \node[anchor=south, inner sep=0] (input) at (0,\nextfigheight) {\includegraphics[width=\figwidth]{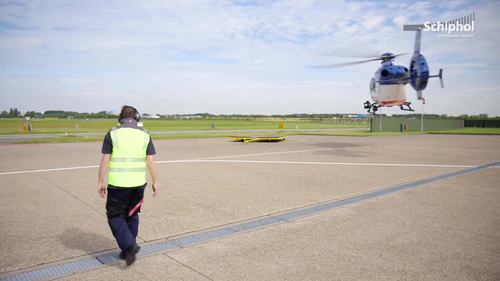}};
\node[anchor=south, inner sep=0] (noisy) at (\figsep,\nextfigheight){\includegraphics[width=\figwidth]{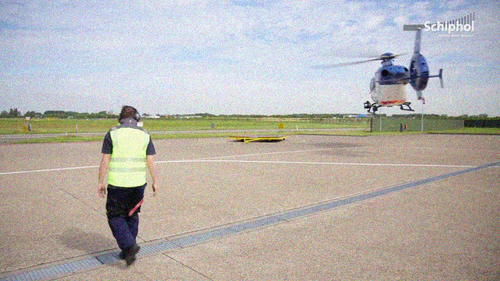}};
\node[anchor=south, inner sep=0] (DNCNN)  at (2*\figsep,\nextfigheight){\includegraphics[width=\figwidth]{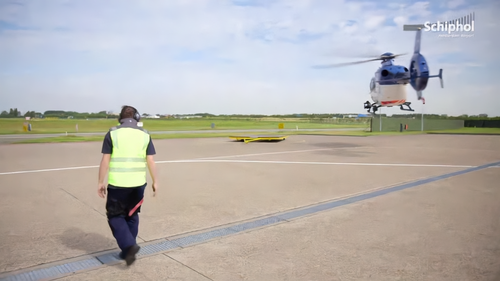}};
\node[anchor=south, inner sep=0] (VBM3D)  at (3*\figsep,\nextfigheight){\includegraphics[width=\figwidth]{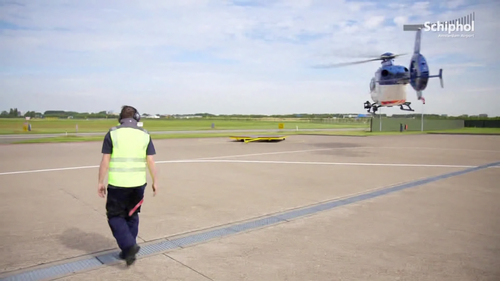}};
\node[anchor=south, inner sep=0] (VNLB)  at (4*\figsep,\nextfigheight){\includegraphics[width=\figwidth]{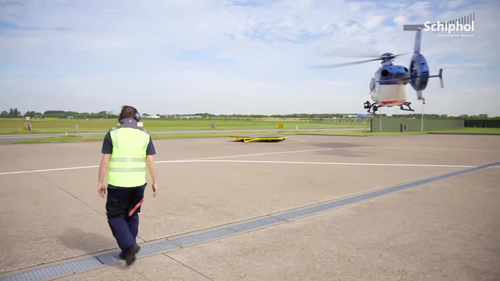}};
\node[anchor=south, inner sep=0] (Ours)  at (5*\figsep,\nextfigheight){\includegraphics[width=\figwidth]{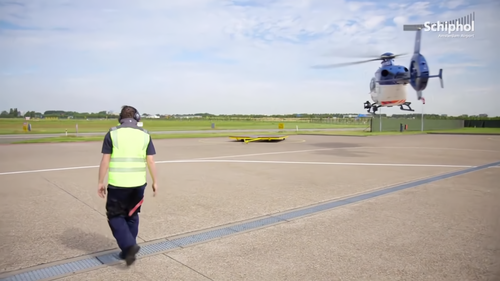}};

    \node [anchor=south] at (input.north) {\scriptsize Input};
    \node [anchor=south] at (noisy.north) {\scriptsize \vphantom{p}Noisy\vphantom{p}};
    \node [anchor=south] at (VBM3D.north) {\scriptsize \vphantom{p}VBM3D\vphantom{p}};
    \node [anchor=south] at (DNCNN.north) {\scriptsize \vphantom{p}DnCNN\vphantom{p}};
    \node [anchor=south] at (VNLB.north) {\scriptsize \vphantom{p}VNLB\vphantom{p}};
    \node [anchor=south] at (Ours.north) {\scriptsize \vphantom{p}VNLnet (Ours)\vphantom{p}};

\addtolength{\nextfigheight}{-0.51\figsep}

\node[anchor=south, inner sep=0] () at (0,\nextfigheight) {\includegraphics[width=\figwidth]{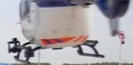}};
\node[anchor=south, inner sep=0] () at (\figsep,\nextfigheight){\includegraphics[width=\figwidth]{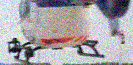}};
\node[anchor=south, inner sep=0] ()  at (2*\figsep,\nextfigheight){\includegraphics[width=\figwidth]{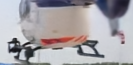}};
\node[anchor=south, inner sep=0] ()  at (3*\figsep,\nextfigheight){\includegraphics[width=\figwidth]{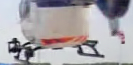}};
\node[anchor=south, inner sep=0] ()  at (4*\figsep,\nextfigheight){\includegraphics[width=\figwidth]{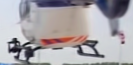}};
\node[anchor=south, inner sep=0] ()  at (5*\figsep,\nextfigheight){\includegraphics[width=\figwidth]{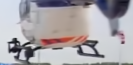}};

\addtolength{\nextfigheight}{-0.53\figsep}

\node[anchor=south, inner sep=0] () at (0,\nextfigheight) {\includegraphics[width=\figwidth]{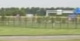}};
\node[anchor=south, inner sep=0] () at (\figsep,\nextfigheight){\includegraphics[width=\figwidth]{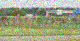}};
\node[anchor=south, inner sep=0] ()  at (2*\figsep,\nextfigheight){\includegraphics[width=\figwidth]{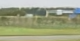}};
\node[anchor=south, inner sep=0] ()  at (3*\figsep,\nextfigheight){\includegraphics[width=\figwidth]{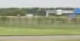}};
\node[anchor=south, inner sep=0] ()  at (4*\figsep,\nextfigheight){\includegraphics[width=\figwidth]{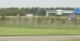}};
\node[anchor=south, inner sep=0] ()  at (5*\figsep,\nextfigheight){\includegraphics[width=\figwidth]{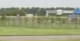}};

\addtolength{\nextfigheight}{-0.50\figsep}

\node[anchor=south, inner sep=0] () at (0,\nextfigheight) {\includegraphics[width=\figwidth]{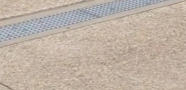}};
\node[anchor=south, inner sep=0] () at (\figsep,\nextfigheight){\includegraphics[width=\figwidth]{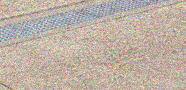}};
\node[anchor=south, inner sep=0] ()  at (2*\figsep,\nextfigheight){\includegraphics[width=\figwidth]{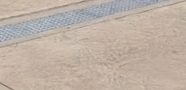}};
\node[anchor=south, inner sep=0] ()  at (3*\figsep,\nextfigheight){\includegraphics[width=\figwidth]{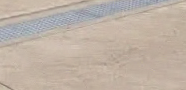}};
\node[anchor=south, inner sep=0] ()  at (4*\figsep,\nextfigheight){\includegraphics[width=\figwidth]{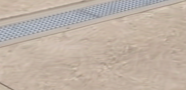}};
\node[anchor=south, inner sep=0] ()  at (5*\figsep,\nextfigheight){\includegraphics[width=\figwidth]{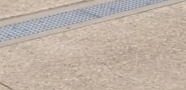}};

\addtolength{\nextfigheight}{-0.51\figsep}

\node[anchor=south, inner sep=0] () at (0,\nextfigheight) {\includegraphics[width=\figwidth]{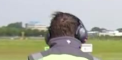}};
\node[anchor=south, inner sep=0] () at (\figsep,\nextfigheight){\includegraphics[width=\figwidth]{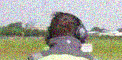}};
\node[anchor=south, inner sep=0] ()  at (2*\figsep,\nextfigheight){\includegraphics[width=\figwidth]{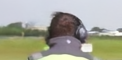}};
\node[anchor=south, inner sep=0] ()  at (3*\figsep,\nextfigheight){\includegraphics[width=\figwidth]{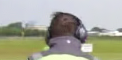}};
\node[anchor=south, inner sep=0] ()  at (4*\figsep,\nextfigheight){\includegraphics[width=\figwidth]{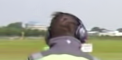}};
\node[anchor=south, inner sep=0] ()  at (5*\figsep,\nextfigheight){\includegraphics[width=\figwidth]{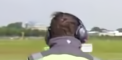}};

 \end{tikzpicture}
 
\caption{Example of denoised result for several algorithms (noise standard deviation of 20) on a sequence of the color DAVIS dataset~\cite{davis}. The crops highlight the results on non-moving and moving parts of the video.}
\label{fig:figFcolor}
\end{figure*}
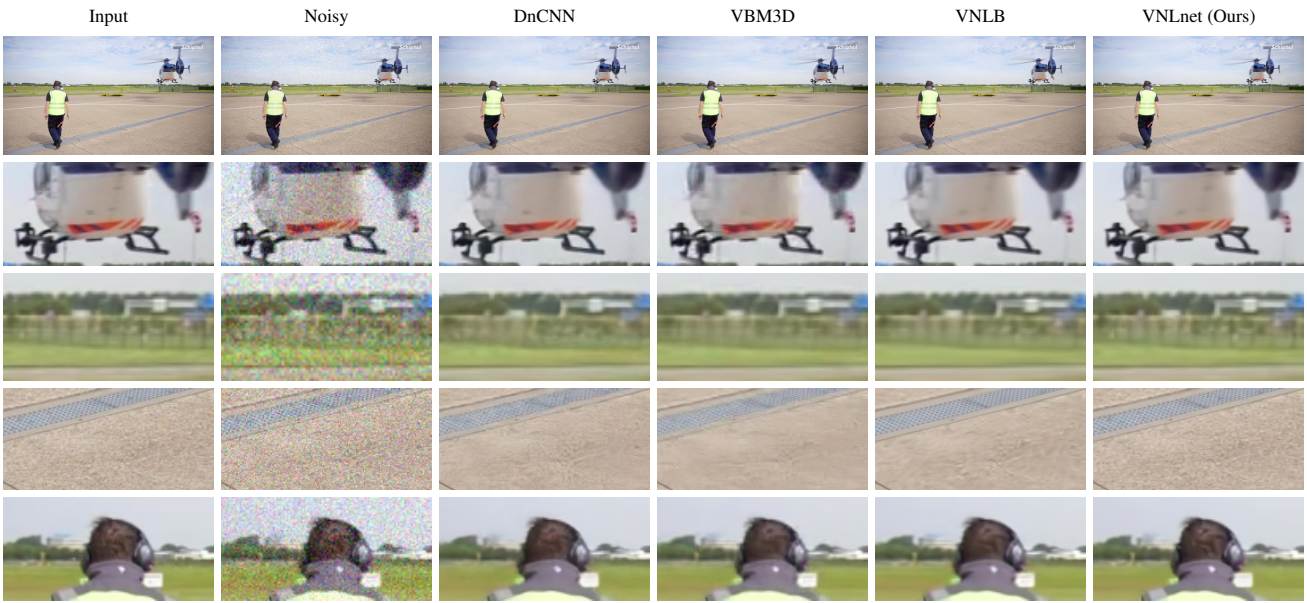

\begin{table}[htp!]
	\begin{center}
		{\small
		\renewcommand{\tabcolsep}{1.0mm}
		\renewcommand{\arraystretch}{1.}
		\begin{tabular}{| l | c | c  |}
		\hline
			Method & Corr. Gaussian noise & Uniform S\&P 25\%              \\\hline
            VNLB & \SB{25.39} / \SB{.5922} & \SB{23.49} / \SB{.7264} \\
            VNLnet & \B{30.94} / \B{.9452} & \B{48.12} / \B{.9951} \\\hline
		\end{tabular}}
	\end{center}
    \caption{Performance (PSNR and SSIM) of VNLB and VNLnet (our method) on the grayscale DERF dataset (Table~\ref{tab:psnr-classic-gray}) for non-standard noises.}
    \label{tab:test_derf_ns}
\end{table}

\begin{table}[htp!]
	\begin{center}
		{\small
		\renewcommand{\tabcolsep}{1.0mm}
		\renewcommand{\arraystretch}{1.}
		\begin{tabular}{| l | c | c | c  |}
			\hline
			Method & $\sigma=10$ & $\sigma=20$ & $\sigma=40$              \\\hline
            VBM3D & 37.43 / .9425 & 33.75 / .8870 & 30.12 / .8068 \\
            VNLB & \SB{38.84} / \SB{.9634} & \SB{35.26} / \SB{.9240} & \SB{31.88} / \SB{.8622}\\
            DnCNN & 36.80 / .9451 & 32.94 / .8878 & 28.69 / .7940 \\
            VNLnet & \B{39.07} / \B{.9663} & \B{35.46} / \B{.9299} & \B{31.90} / \B{.8659} \\\hline
		\end{tabular}}
	\end{center}
    \caption{Performance (PSNR and SSIM) of DnCNN, VBM3D and VNLnet (our method) on the grayscale DAVIS dataset~\cite{davis} for several noise levels $\sigma$ (10, 20 and 40).}
    \label{tab:test_davis}
\end{table}

\begin{table}[htp!]
	\begin{center}
		{\small
		\renewcommand{\tabcolsep}{1.0mm}
		\renewcommand{\arraystretch}{1.}
		\begin{tabular}{| l | c | c | c  |}
			\hline
			Method & $\sigma=10$ & $\sigma=20$ & $\sigma=40$              \\\hline
            VBM3D & 38.43 / .9591  &  34.74 / .9157  &  31.38 / .8473  \\
            VNLB &  \SB{40.31} / \SB{.9725}  &  \SB{36.79} / \SB{.9420}  & \SB{33.34} / \SB{.8896} \\
            DnCNN &  38.91 / .9655 &  35.24 / .9278  &  31.81 / .8637  \\
            VNLnet &  \B{40.71} / \B{.9760}  &  \B{37.39} / \B{.9534}  &  \B{33.96} / \B{.9091}  \\\hline
		\end{tabular}}
	\end{center}
    \caption{Performance (PSNR and SSIM) of DnCNN, VBM3D and VNLnet (our method) on the color DAVIS dataset~\cite{davis} for several noise levels $\sigma$ (10, 20 and 40).}
    \label{tab:test_davis_rgb}
\end{table}

\begin{figure}
\centering

\centering
\begin{tikzpicture}
  \setlength{\nextfigheight}{0cm}
  \setlength{\figwidth}{.24\columnwidth}
  \setlength{\figsep}{.25\columnwidth}

\node[anchor=south, inner sep=0] (input) at (0*\figsep,\nextfigheight){\includegraphics[width=\figwidth]{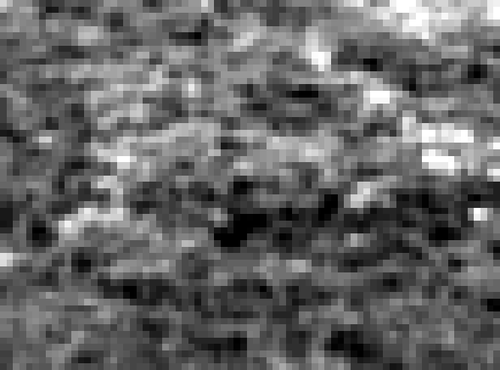}};
\node[anchor=south, inner sep=0] (noisy) at (1*\figsep,\nextfigheight){\includegraphics[width=\figwidth]{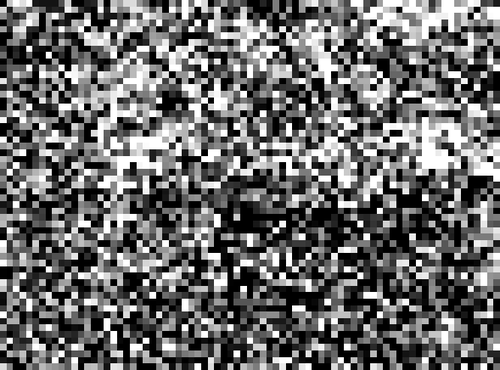}};
\node[anchor=south, inner sep=0] (Ours)  at (3*\figsep,\nextfigheight){\includegraphics[width=\figwidth]{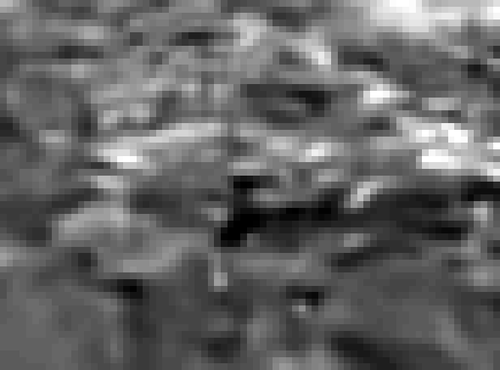}};
\node[anchor=south, inner sep=0] (VNLB)  at (2*\figsep,\nextfigheight){\includegraphics[width=\figwidth]{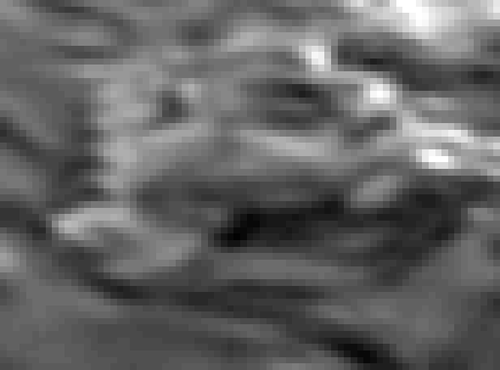}};

\node [anchor=south] at (input.north) {\scriptsize Input};
\node [anchor=south] at (noisy.north) {\scriptsize \vphantom{p}Noisy\vphantom{p}};
\node [anchor=south] at (Ours.north) {\scriptsize \vphantom{p}VNLnet (Ours)\vphantom{p}};
\node [anchor=south] at (VNLB.north) {\scriptsize \vphantom{p}VNLB\vphantom{p}};

\addtolength{\nextfigheight}{-0.82\figsep}

\node[anchor=south, inner sep=0] () at (0,\nextfigheight) {\includegraphics[width=\figwidth]{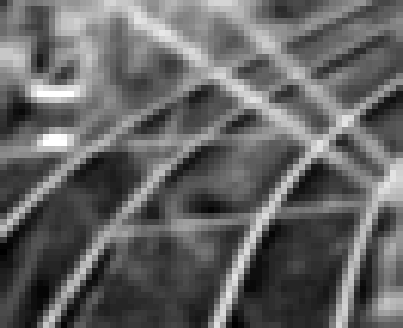}};
\node[anchor=south, inner sep=0] () at (\figsep,\nextfigheight){\includegraphics[width=\figwidth]{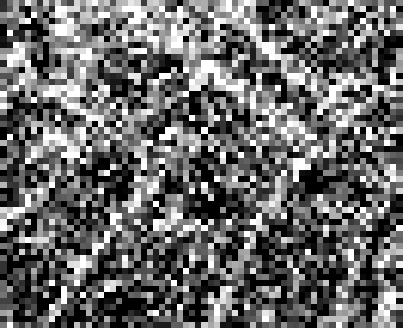}};
\node[anchor=south, inner sep=0] ()  at (2*\figsep,\nextfigheight){\includegraphics[width=\figwidth]{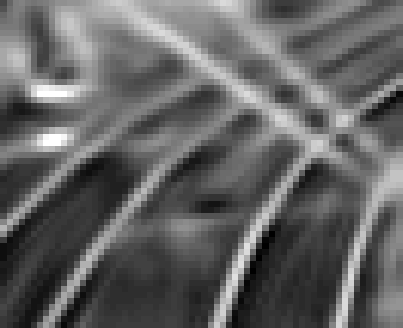}};
\node[anchor=south, inner sep=0] ()  at (3*\figsep,\nextfigheight){\includegraphics[width=\figwidth]{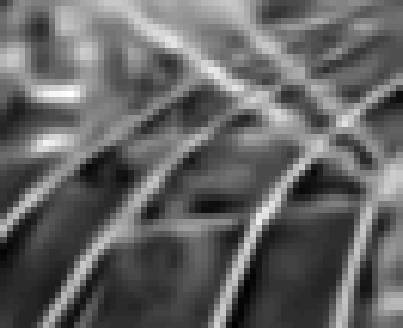}};

 \end{tikzpicture}

\caption{Examples of background details more properly recovered by VNLnet compared to VNLB (enhanced contrast).}
\label{fig:vnlb_vs_vnlnet}
\end{figure}
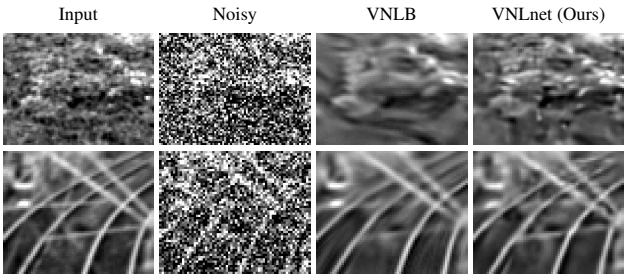

\begin{figure*}
\centering

\begin{tikzpicture}
  \setlength{\nextfigheight}{0cm}
  \setlength{\figwidth}{0.28\textwidth}
  \setlength{\figsep}{0.285\textwidth}
    \node[anchor=west, inner sep=0] (input) at (0,\nextfigheight) {\includegraphics[width=\figwidth]{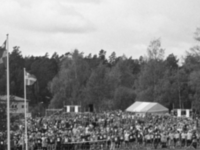}};
\node[anchor=west, inner sep=0] () at (\figsep,\nextfigheight){\includegraphics[width=\figwidth]{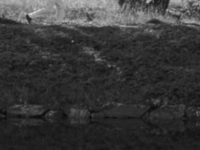}};
\node[anchor=west, inner sep=0] ()  at (2*\figsep,\nextfigheight){\includegraphics[width=\figwidth]{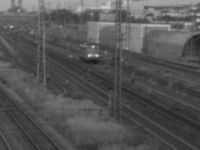}};

\node [anchor=east] at (input.west) {\scriptsize Input};

\addtolength{\nextfigheight}{-0.75\figsep}

\node[anchor=west, inner sep=0] (noisy) at (0,\nextfigheight) {\includegraphics[width=\figwidth]{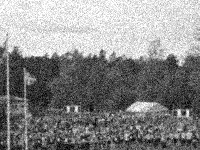}};
\node[anchor=west, inner sep=0] () at (\figsep,\nextfigheight){\includegraphics[width=\figwidth]{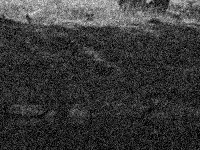}};
\node[anchor=west, inner sep=0] ()  at (2*\figsep,\nextfigheight){\includegraphics[width=\figwidth]{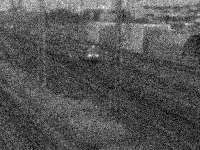}};

\node [anchor=east] at (noisy.west) {\scriptsize Noisy};

\addtolength{\nextfigheight}{-0.75\figsep}

\node[anchor=west, inner sep=0] (DnCNN) at (0,\nextfigheight) {\includegraphics[width=\figwidth]{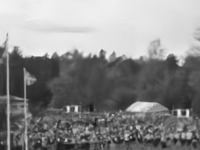}};
\node[anchor=west, inner sep=0] () at (\figsep,\nextfigheight){\includegraphics[width=\figwidth]{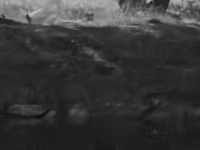}};
\node[anchor=west, inner sep=0] ()  at (2*\figsep,\nextfigheight){\includegraphics[width=\figwidth]{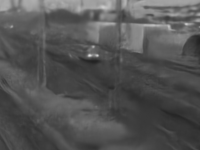}};

\node [anchor=east] at (DnCNN.west) {\scriptsize DnCNN};

\addtolength{\nextfigheight}{-0.75\figsep}

\node[anchor=west, inner sep=0] (VBM3D) at (0,\nextfigheight) {\includegraphics[width=\figwidth]{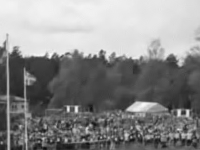}};
\node[anchor=west, inner sep=0] () at (\figsep,\nextfigheight){\includegraphics[width=\figwidth]{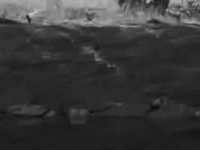}};
\node[anchor=west, inner sep=0] ()  at (2*\figsep,\nextfigheight){\includegraphics[width=\figwidth]{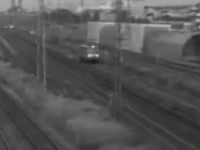}};

\node [anchor=east] at (VBM3D.west) {\scriptsize VBM3D};

\addtolength{\nextfigheight}{-0.75\figsep}

\node[anchor=west, inner sep=0] (VNLB) at (0,\nextfigheight) {\includegraphics[width=\figwidth]{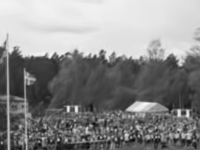}};
\node[anchor=west, inner sep=0] () at (\figsep,\nextfigheight){\includegraphics[width=\figwidth]{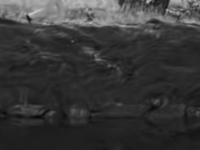}};
\node[anchor=west, inner sep=0] ()  at (2*\figsep,\nextfigheight){\includegraphics[width=\figwidth]{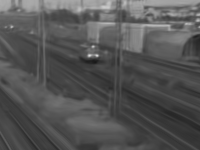}};

\node [anchor=east] at (VNLB.west) {\scriptsize VNLB};

\addtolength{\nextfigheight}{-0.75\figsep}

\node[anchor=west, inner sep=0] (VNLNET) at (0,\nextfigheight) {\includegraphics[width=\figwidth]{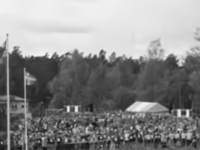}};
\node[anchor=west, inner sep=0] () at (\figsep,\nextfigheight){\includegraphics[width=\figwidth]{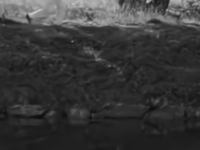}};
\node[anchor=west, inner sep=0] ()  at (2*\figsep,\nextfigheight){\includegraphics[width=\figwidth]{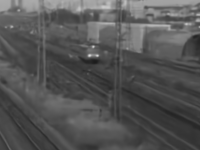}};

\node [anchor=east,align=left] at (VNLNET.west) {\scriptsize \parbox{3em}{VNLnet (Ours)}};%

 \end{tikzpicture}
\caption{Examples of areas where the  level of restored detail of the methods differs significantly (noise standard deviation of 20)  on \textit{crowd, park} and \textit{station}.}
\label{fig:examples}
\end{figure*}

On tables \ref{tab:psnr-classic-gray} and \ref{tab:psnr-classic-color}, we show a comparison of DnCNN (applied frame-by-frame) and the proposed method \textit{Video Non-Local Network} (VNLnet) with other state-of-the-art video denoising methods~\cite{arias2018comparison} for the DERF dataset. The state-of-the-art methods include SPTWO~\cite{buades2016patch}, VBM3D~\cite{Dabov2007v}, VBM4D~\cite{Maggioni2011} and 
VNLB~\cite{Arias2015}. Figures \ref{fig:figA}, \ref{fig:examples} and \ref{fig:vnlb_vs_vnlnet} show results for
some of the most relevant methods.

DnCNN, VBM3D and VNLB were also compared on the DAVIS \texttt{test-dev} dataset~\cite{davis}. Results are shown in Tables \ref{tab:test_davis}, \ref{tab:test_davis_rgb} and Figure~\ref{fig:figFcolor}.
VNLnet is the best performing method in the DAVIS dataset, but it is outperformed by VNLB (Video Non-Local Bayes) on the DERF dataset. This could be explained by the fact that the DERF dataset is blurrier compared to DAVIS and to VNLnet's training set. This blur is caused by a wider antialiasing filter used when downscaling the original videos in~\cite{arias2018comparison}. We notice DnCNN also underperforms on this dataset compared to the sharper DAVIS dataset.

A comparison of tables \ref{tab:test_davis} and \ref{tab:test_davis_rgb} reveals that CNN based methods are better in exploiting the correlations between color channels: while for grayscale, VBM3D was significantly outperforming DnCNN in PSNR on the DAVIS dataset, the reverse occurs for color. In addition, the gap between VNLnet and VNLB is widened in color. This should not come as a surprise, since the way in which VBM3D and VNLB treat color is rather heuristic: an orthogonal color transformed is applied to the video which is supposed to decorrelate color information. Based on this, the processing of each color channel of a group of patches in done independently.
These results make VNLnet the state-of-the-art method for video denoising and the first so of the neural kind.

In order to better compare qualitative aspects of the results we show some details in Figures~\ref{fig:figA},~\ref{fig:examples} and~\ref{fig:vnlb_vs_vnlnet}. Figure~\ref{fig:figA} shows some results on an video of DERF. The two bottom rows show details on two different types of areas (background and moving object). We include as reference the \textit{Non-Local Pixel Mean}, which is just the result of the averaging of the matches presented to the network. As noise remains, one can thus see that the network 
does more than averaging the data on static areas (last two rows). On background objects, the denoising performance is significantly improved compared to DnCNN and is similar to VNLB (middle row). For some scenes, VNLnet recovers significantly more details in the background, as shown on Figure~\ref{fig:vnlb_vs_vnlnet} and Figure~\ref{fig:examples}. On moving foreground objects - 
thus with bad matches - our method performs similar to DnCNN, as can be observed on the last row of Figure~\ref{fig:figA}. In general, we observe VNLnet has better background reconstruction than VNLB. Both methods achieve high temporal consistency, which is an important quality requirement for video denoising.

One of the benefits of CNNs over traditional model-based approaches is that they can be easily retargeted to handle other noise models. To illustrate this we compare the classical best performing method VNLB and our method on non-standard noises on Table~\ref{tab:test_derf_ns}. As expected, VNLnet significantly beats VNLB for these non-Gaussian noise distributions. 

\paragraph{A note on running times.}

\begin{table}%
	
	\begin{center}
		{\small
		\renewcommand{\tabcolsep}{1.0mm}
		\renewcommand{\arraystretch}{1.}
		\begin{tabular}{| c | c | c | c | c |}
			\hline
        VBM3D & DnCNN & VBM4D & VNLB & SPTWO\\\hline
        1.3s & 13s & 52s & 140s & 210s
			  \\\hline
		\end{tabular}}
	\end{center}
    \caption{Running time per frame on a $960 \times 540$ video for VBM3D, DnCNN, VBM4D, VNLB and SPTWO on single CPU core.}%
    \label{tab:cpu_speed}
\end{table}

\begin{table}[htp!]
	\begin{center}
		{\small
		\renewcommand{\tabcolsep}{1.0mm}
		\renewcommand{\arraystretch}{1.}
		\begin{tabular}{| c | c | c |}
			\hline
			Non-local search & Rest of the network & DnCNN             \\\hline
            850 ms & 80 ms & 95 ms
			  \\\hline
		\end{tabular}}
	\end{center}
    \caption{Running time per frame on a $960 \times 540$ video on a NVIDIA TITAN V ($41\times41$ patches at every position, $41\times41\times15$ 3D windows, the default parameters).}
    \label{tab:titan_speed}
\end{table}

On Table \ref{tab:cpu_speed}, we compare the CPU running time of VBM3D, DnCNN and VNLB when denoising a video frame. While we do not a have a CPU implementation of the patch search layer, the GPU runtimes of Table \ref{tab:titan_speed} point out that on CPU our method should be 10 times slower than DnCNN. The non-local search is particularly costly because we search matches on 15 frames for patches centered in every pixel of our image. The patch search  could be made significantly faster by reducing the size of the 3D window using tricks explored in other papers.   VBM3D for  example centers the  search on each frame on small windows around the best matches found  in the previous frame.  A related acceleration is to use a search strategy based on  PatchMatch~\cite{Barnes:2009:patchmatch}.

\section{Implementation details}

The patch search requires the computation of the distance between each patch in the image and the patches in the search region. If implemented na\"ively, this operation can be prohibitive. Patch-based methods require a patch search step. To reduce the computational cost, a common approach is to search the nearest neighbors only for the patches in a subgrid of the image. For example BM3D processes 1/9th of the patches with default parameters. Since the processed patches overlap, the aggregation of the denoised patches covers the whole image.

Our proposed method does not have any aggregation. We compute the neighbors for all image patches, which is costly. In the case of video, best results are obtained with large patches and a large search region (both temporally and spatially). Therefore we need a highly efficient patch search algorithm.

Our implementation uses an optimized GPU kernel which searches for the locations in parallel. For each patch, the best distances with respect to all other patches in the search volume are maintained in a table. We split the computation of the distances is two steps: first compute the sum of squares across columns:
\[D^{\text{col}}(x',y',t') = \sum_{h=-s/2}^{s/2} (v(x,y+h,t) - v(x', y'+h,t')^2.\]
Then the distances can be obtained by applying a horizontal box filter of size $s$ on the volume $D^{\text{col}}$ composed by the neighboring GPU threads. The resulting implementation has linear complexity in the size of the search region and the patch width. 

To optimize the speed of the algorithm we use the GPU shared memory as cache for the memory accesses thus reducing bandwidth limitations. In addition, for sorting the distances the ordered table is stored into GPU registers, and written to memory only at the end of the computation. The computation of the L2 distances and the maintenance of the ordered table have about the same order of computation cost. More details about the implementation can be found in Appendix \ref{sec:more-implementation-details}.

\section{Conclusions}

We described a simple yet effective way of incorporating non-local information into a CNN for video denoising. The proposed method computes for each image patch the $n$ most similar neighbors on a spatio-temporal window and gathers the value of the central pixel of each similar patch to form a non-local feature vector which is given to a CNN. Our method 
yields a significant gain compared to using the single frame baseline CNN on each video frame. Compared to other video denoising algorithms, it achieves state-of-the-art performance and attaining the highest PSNR on a downscaled version of the DAVIS dataset.

Our contribution places neural networks among the best video denoising methods and opens the way for new works in this area. 

We have seen the importance of having reliable matches: On the validation set, the best performing method used patches of size $41\times41$ for the patch search. We have also noticed that on regions with non-reliable matches (complex motion), the network reverts to a result similar to single image denoising. Thus we believe future works should focus on improving this area, by possibly adapting the size of the patch and passing information about the quality of the matches to the network.

\appendix

\section{More implementation details on the non-local search}
\label{sec:more-implementation-details}

In this section we describe in more details our GPU implementation of the non-local search.

As mentioned in the main paper, a na\"ive implementation of the patch search can be very inefficient.

The patch search algorithm can be divided conceptually into two parts: First, computing for all positions in the search window the L2 distance between the reference patch and the target patch, both of size $K\times K$. Second, retaining the best $N$ distances and positions. Both parts need to be implemented efficiently.

\begin{algorithm}%
	\SetAlgoLined
	\DontPrintSemicolon
    {\bf Input:} New position $p$ and distance $d$\;
    {\bf Input:} Tables Positions and Distances of length $N$.\;

    \If{d $<$ Distances[N-1]}{
        \For{i from N-1 to 1}{
            insert $\leftarrow$ Distances[i-1] $\le$ d\;
				Positions[i] $\leftarrow$ p \textbf{if} insert \textbf{else} Positions[i-1]\;
				Distances[i] $\leftarrow$ d \textbf{if} insert \textbf{else} Distances[i-1]\;
				quit function \textbf{if} insert
        }
        Positions[0] $\leftarrow$ p\;
        Distances[0] $\leftarrow$ d\;
    }
	\label{algo:table}
    \caption{Keeping an ordered table of distances and positions}
\end{algorithm}

Of the two parts, the most critical one is the maintenance of the table of the best $N$ distances and positions. Our implementation maintains the distances and positions in an ordered table stored into the GPU registers. Indeed GPUs have many available registers: a maximum of 128 on Intel, and 256 on AMD and NVIDIA for the current generation, although consuming fewer registers helps reducing the latency. Thus if $N$ is small enough, both tables (distances and positions) can be stored in the register table. The tables need to be accessed very frequently, so not using registers leads to a much slower code. Our algorithm is summarized on Algorithm \ref{algo:table}.

\begin{algorithm}%
	\SetAlgoLined
	\DontPrintSemicolon
    On CPU: Divide the image into regions of overlapping row segments of length 128. The overlap should be of length patch\_width - 1.\;
    On GPU:\;
    Assign one thread to each element of an horizontal segment.\;
    \For{each offset (dz, dy, dx) in the 3D search window}{
    Compute squared L2 distance between reference and target center columns\;
    Write result into GPU shared memory\;
    Sum results of neighboring threads\;
    Maintain table of best distances and positions\;
    }
    Only save valid results (border threads can't compute the full distance)\;
	\label{algo:nlsearch}
    \caption{Summary of our patch search implementation}
\end{algorithm}

Then comes the computation of the L2 distances between  patches. A na\"ive algorithm would, for every patch pair, read the reference patch, read the target patch, and compare them pixel-wise, without reusing any computation or memory access. Our optimized algorithm uses the fact that the L2 distances computation share a lot of common elements with the same computations after a translation of the positions of the reference and the target patches. This avoids both computation and memory accesses. We organize GPU threads into groups treating an horizontal segment each. Each thread will compute the distance between the reference and the target patch for the center column only, and shared GPU memory will be used to read the results of neighboring threads and compute the full distance. Since we only need to compare a column of the patches, that column can be stored into GPU registers, thus avoiding to reload the reference patch data every iteration. Threads at the border of the segments can't compute the full distance as some results are missing, thus some overlap between the segments are required. We found a length of 128 to be a good compromise for the length of these segments.  For increased speed, we cache our memory accesses with the GPU shared memory between the computing threads. The process is summarized in Algorithm \ref{algo:nlsearch}.

Both our implementation and the na\"ive implementation have a linear complexity in the size of the 3D search window, but our algorithm has a linear complexity with respect to the width of the patches, while it is quadratic for the na\"ive algorithm. One should know, if not familiar with patch search, that the 3D search window defines the search region for all patches whose center lie inside the 3D search window, thus the patches do not have to fit completely inside the region. For the default VNLnet parameters, our implementation is $25$ times faster (on a NVIDIA TITAN V) than the na\"ive implementation (both using Algorithm \ref{algo:table} for the tables of distances and positions).

{\small
\bibliographystyle{ieee}
\bibliography{image_denoising,video_denoising,deep-denoising}
}

\end{document}